\documentclass[10pt,twocolumn,letterpaper]{article}

\usepackage{cvpr}              %

\usepackage[dvipsnames]{xcolor}

\definecolor{cvprblue}{rgb}{0.21,0.49,0.74}
\usepackage[pagebackref,breaklinks,colorlinks,citecolor=cvprblue]{hyperref}
\usepackage{booktabs}
\usepackage{multirow}
\usepackage{makecell}

\def\paperID{1090} %
\def\confName{CVPR}
\def\confYear{2024}

\title{IPT-V2: Efficient Image Processing Transformer using Hierarchical Attentions}

\author{Zhijun Tu\footnotemark[1], Kunpeng Du\footnotemark[1], Hanting Chen, Hailing Wang, Wei Li, Jie Hu, Yunhe Wang\\
Huawei Noah’s Ark Lab\\
{\tt\small \{zhijun.tu, dukunpeng, chenhanting, wanghailing6, wei.lee, hujie23, yunhe.wang\}@huawei.com}
}

\begin{document}
\maketitle
\footnotetext[1]{Equal contribution}

\begin{abstract}
Recent advances have demonstrated the powerful capability of transformer architecture in image restoration. 
However, our analysis indicates that existing transformer-based methods can not establish both exact global and local dependencies simultaneously, which are much critical to restore the details and missing content of degraded images. 
To this end, we present an efficient image processing transformer architecture with hierarchical attentions, called IPT-V2, adopting a focal context self-attention (FCSA) and a global grid self-attention (GGSA) to obtain adequate token interactions in local and global receptive fields.  
Specifically, FCSA applies the shifted window mechanism into the channel self-attention, helps capture the local context and mutual interaction across channels. And GGSA constructs long-range dependencies in the cross-window grid, aggregates global information in spatial dimension.
Moreover, we introduce structural re-parameterization technique to feed-forward network to further improve the model capability.
Extensive experiments demonstrate that our proposed IPT-V2 achieves state-of-the-art results on various image processing tasks, covering denoising, deblurring, deraining and obtains much better trade-off for performance and computational complexity than previous methods. Besides, we extend our method to image generation as latent diffusion backbone, and significantly outperforms DiTs.
\end{abstract}
    
\section{Introduction}
Image restoration aims to reconstruct the high-quality (HQ) clean images from the low-quality (LQ) degraded ones. such as denoising~\cite{zhang2017beyond,guo2019toward,chang2020spatial}, deblurring~\cite{gopro2017,son2021single,lee2021iterative} and deraining~\cite{li2018recurrent,ren2019progressive,mspfn2020}, \textit{etc}. Due to the ill-posed nature, it is much challenging to distinguish textures and unpleasant degradation, and restore content from LQ images that has been completely lost during degradation processes. With large scale paired datasets, convolutional neural networks (CNNs) show great representation capability to learn the non-linear mapping of LQ to the corresponding HQ. However, CNNs could only capture local features with limited kernel size, while global information is also much critical to restore details because they could provide additional references for the lost content. In contrast, the transformer architecture~\cite{chen2021pre} is well known for modeling long-range dependencies across tokens with multi-head self-attention (SA), but the main drawback is the computational complexity that is quadratic with the resolution. 
\begin{figure}[t]
	\centering
	\includegraphics[width=1.0\linewidth]{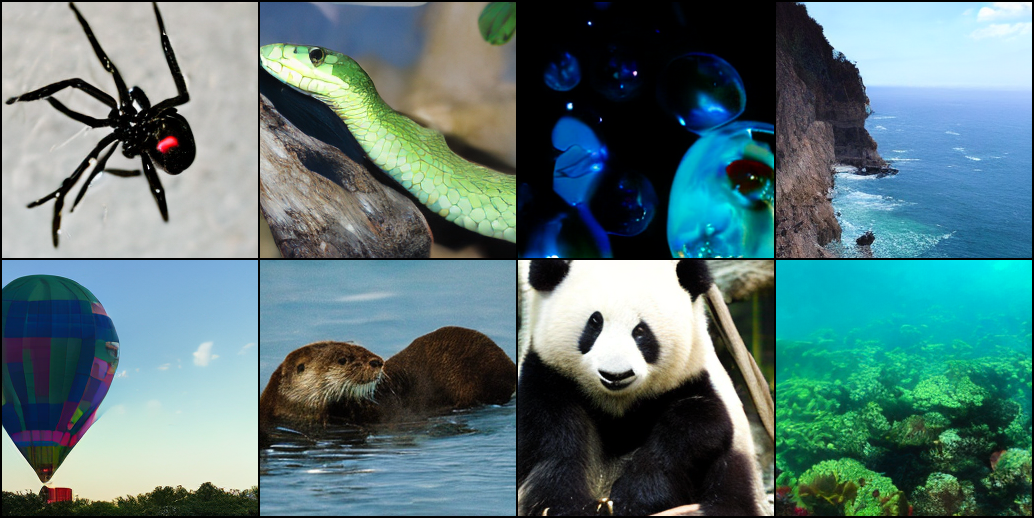}
	\caption{Uncurated generated images by latent IPT-V2 on ImageNet 256$\times$256 dataset~\cite{krizhevsky2012imagenet}.}
	\label{Generation}
\end{figure}
\begin{figure}[t]
	\centering
	\includegraphics[width=1.0\linewidth]{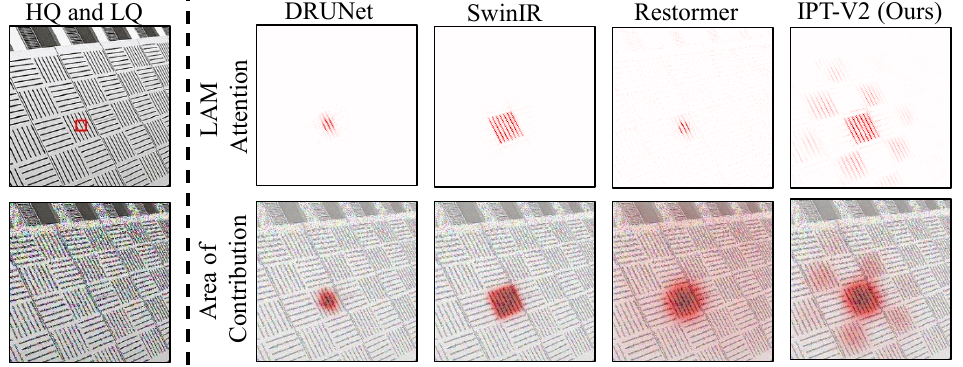}
	\caption{LAM~\cite{gu2021interpreting} results for different architectures. The LAM represents the contributions of each pixel in the input LQ image when restoring the corresponding clear region in the red box. The results shows that IPT-V2 could provide accurate local and global token interactions simultaneously.}
	\label{Motivation}
	\vspace{-5mm}
\end{figure}

\begin{figure*}
	\centering
	\begin{minipage}{0.245\textwidth}
		\centering
		\includegraphics[width=1.0\linewidth]{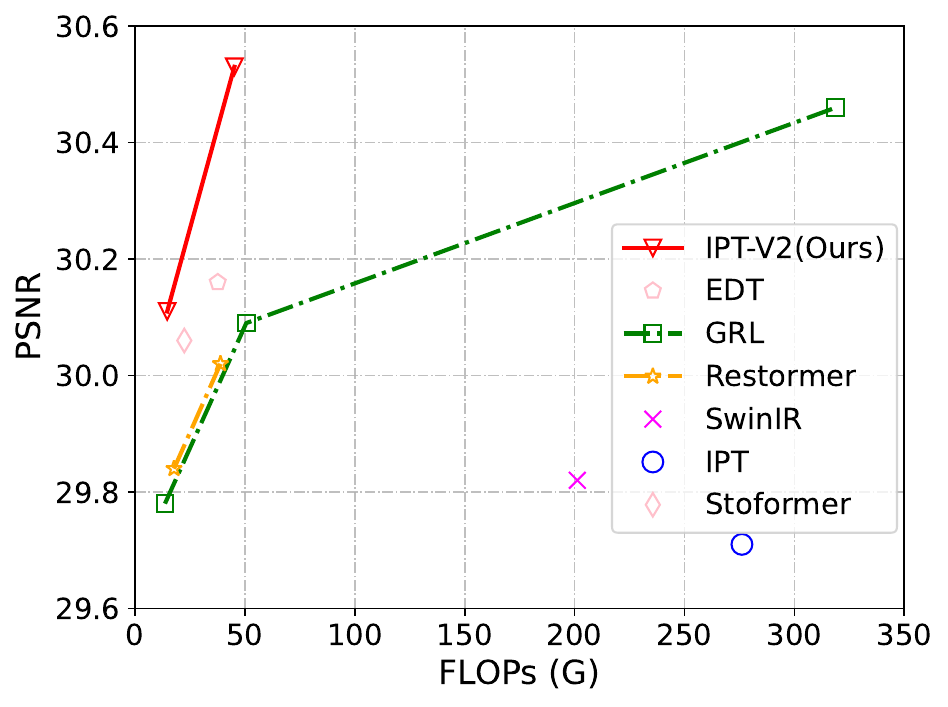}
		\subcaption{Gaussian Denoising}
		\label{denoising}
	\end{minipage}
	\begin{minipage}{0.245\textwidth}
		\centering
		\includegraphics[width=1.0\linewidth]{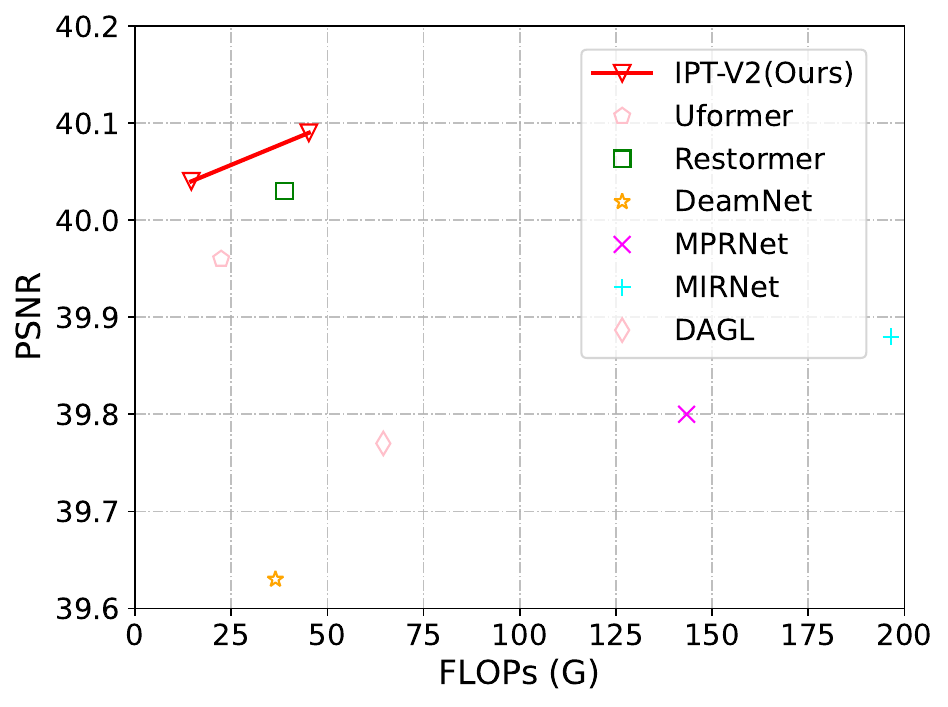}
		\subcaption{Real Denoising}
		\label{vis_12}
	\end{minipage}
	\begin{minipage}{0.245\textwidth}
		\centering
		\includegraphics[width=1.0\linewidth]{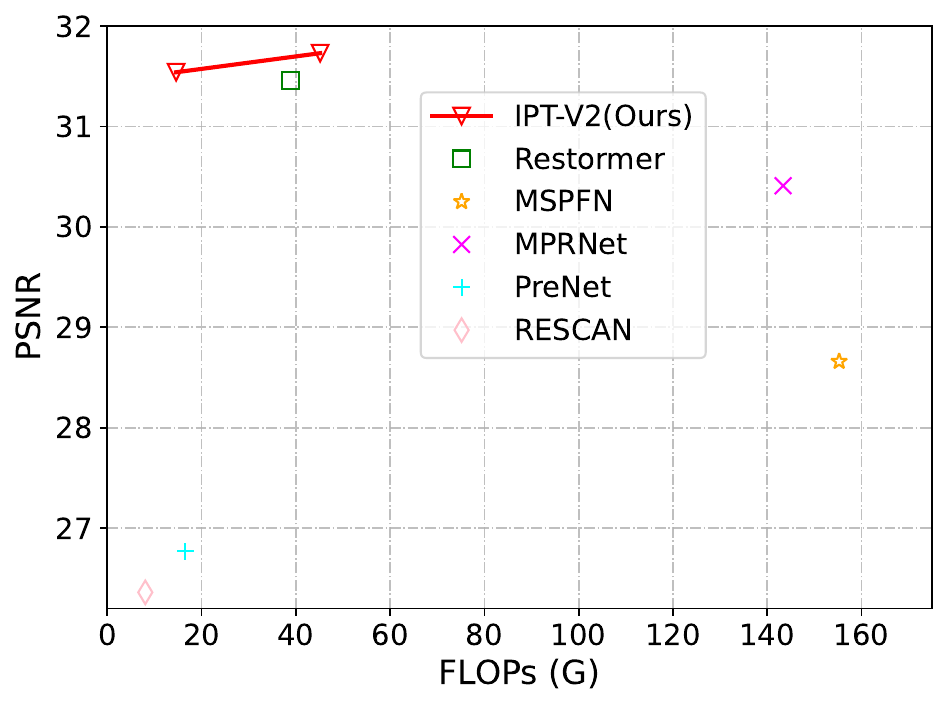}
		\subcaption{Deraining}
		\label{vis_13}
	\end{minipage}
	\begin{minipage}{0.245\textwidth}
		\centering
		\includegraphics[width=1.0\linewidth]{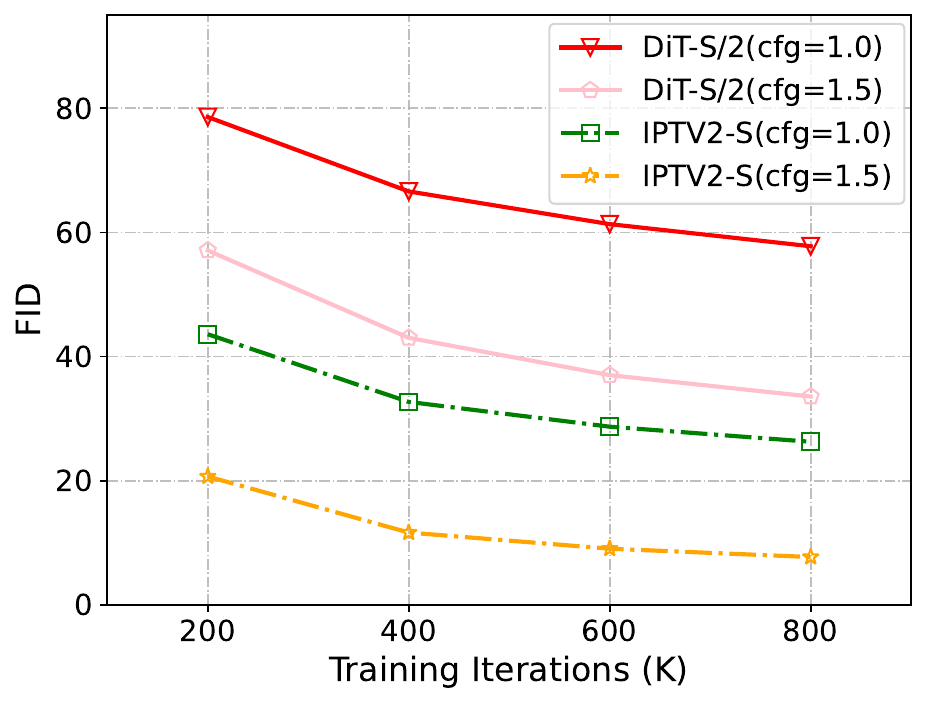}
		\subcaption{Image Generation}
		\label{vis_14}
	\end{minipage}
	\caption{(a-c) PSNR (dB) vs. FLOPS (G) on different image restoration tasks. The results are test on Urban100~\cite{huang2015single}, DND~\cite{plotz2017benchmarking} and Rain100H~\cite{yang2017deep} for Gaussian denoising, real denoising and deraining, respectively. The proposed IPT-V2 achieves state-of-the-art performance on various image restoration tasks and obtains better trade-off for accuracy and computational overhead. (d) Comparison of different training iterations and classifier-free guidance scales on image generation of ImageNet 256$\times$256.}
	\label{PSNR_FLOPS}
\end{figure*}

To balance the complexity and accuracy, most previous methods~\cite{chen2205activating, chen2022cross,li2023efficient,liang2021swinir,xiao2022stochastic} tend to adopt the shifted window self-attention (WSA)~\cite{liu2021swin}, but attention maps are only computed in non-overlapping and fixed-size windows and the shift mechanism can not capture complete cross-window spatial dependencies explicitly. Different from the spatial self-attention, the channel self-attention (CSA)~\cite{zamir2022restormer} is also explored to improve the efficiency of transformer models in image restoration. CSA calculates the self-attention across channels, the computational complexity is linearly related to the resolution. Although CSA could provide global context information, it is much rough for that CSA do not construct the dependencies across pixels in spatial dimension but aggregates them in an average style.

In this paper, we propose to construct accurate local and global interactions with hierarchical attentions in image restoration, and present an efficient Image Processing Transformer architecture, named IPT-V2, it contains three core components: 
(1) \textit{Focal Context Self-Attention (FCSA)}. To obtain adequate  local context  with less computational complexity, we apply channel self-attention into local regions. By sharing the Query-Key-Value projection module, we simultaneously calculate the channel self-attention in the window and shifted window and merge them with learnable spatial weight, which also helps compensate for the edge effect caused by non-overlapping window. 
(2) \textit{Global Grid Self-Attention (GGSA)}. Different from the spatial self-attention in vanilla ViT~\cite{dosovitskiy2020image}, 
we propose to partition the feature map into an uniform grid, and calculate self-attention of pixels with same position in each cubby, which could establish truly long-range dependencies and aggregate global information to each cubby with much less computational overhead. 
(3) \textit{Re-Parameterization Locally-enhance Feed-Forward Network (Rep-LeFFN)}.To further enhance the capability of IPT-V2, we apply the sequential and parallel structural re-parameterization techniques to the locally-enhance feed-forward network~\cite{wang2022uformer} during training, and remain the original structure in the inference stage.

Furthermore, to demonstrate the effectiveness of our proposed IPT-V2, we conduct experiments on various image restoration tasks, including real-world denoising, synthetic denoising with Gaussian grayscale and color noise, single image motion deblurring, defocus deblurring (single and dual tasks) and image deraining. Figure~\ref{PSNR_FLOPS} shows that our proposed IPT-V2 not only achieves the state-of-the-arts performance, but also obtains much better trade-off for accuracy and complexity than previous architectures. Besides, we also apply IPT-V2 in image generation, replacing DiTs~\cite{peebles2023scalable} in the latent space, significantly boot the performance, as shown in Figure~\ref{Generation}. 

The contributions are summarized as follow:

(1) We present IPT-V2, a novel transformer-based architecture with hierarchical attentions for image restoration and generation, which is both efficient and effective.

(2) The proposed focal context self-attention, global grid self-attention and re-parameterization locally enhance feed-forward network could construct accurate token interactions in local and global range explicitly for image restoration.

(3) Extensive experiments demonstrate that IPT-V2 achieves state-of-the-art performance on various image restoration datasets and benchmarks with less complexity, including denoising, deblurring, deraining, and shows excellent performance on image generation.

\section{Related Works}
\textbf{Convolution Neural Networks.}
For the past decade, CNNs~\cite{dong2014learning,kim2016accurate,zhang2017learning,zhang2018image,zhang2020residual} have achieved superior performance in image restoration due to their powerful representation capability. With large scale paired datasets, deep-learning based methods could learn the non-linear mapping from low-quality images to high-quality images. They explore the different neural network structures (valinna shape~\cite{kim2016accurate}, U-shape~\cite{zamir2021multi} and multi-branch~\cite{lim2017enhanced, zhang2020residual}) or attention mechanism (channel attention and spatial attention~\cite{lim2017enhanced,li2018recurrent,wang2020eca,zamir2020learning,zamir2021multi}, non-local attention~\cite{liu2018non}), greatly accelerating the development of image restoration. We
refer the reader to NTIRE challenge reports~\cite{ignatov2019ntire, abuolaim2021ntire} for details.
\begin{figure*}[t]
	\centering
	\includegraphics[width=1.0\linewidth]{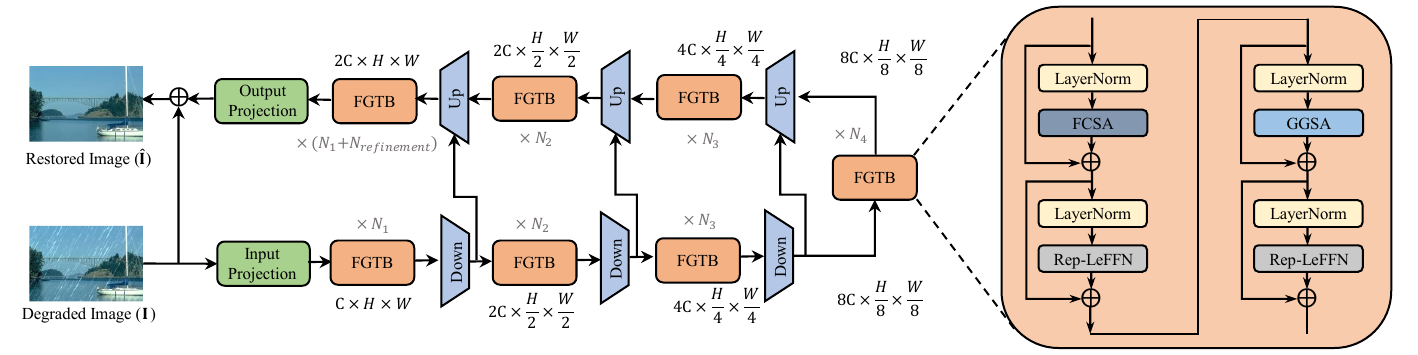}
	\caption{The overview of IPT-V2 architecture and focal and global transformer block (FGTB), consisting of focal context self-attention (FCSA), global grid self-attention (GGSA) and Re-parameterization LeFFN (Rep-LeFFN).}
	\label{IPT-V2}
\end{figure*}

\noindent \textbf{Vision Transformer.}
The transformer architecture~\cite{vaswani2017attention} was first introduced in natural language processing. Due to its more superior capacity compared to CNNs, it has been extensively applied in image restoration in recent years. 
IPT~\cite{chen2021pre} has pioneered the application of the standard vision transformer architecture to image restoration, which achieves outstanding performance with pretrained on ImageNet dataset. SwinIR~\cite{liang2021swinir} employed the swin transformer layers to achieve high-quality image restoration for low-resolution inputs. Inspired by U-Net~\cite{ronneberger2015u}, Uformer~\cite{wang2022uformer} proposed a U-shape transformer architecture for image restoration. Restormer~\cite{zamir2022restormer} built an efficient model by making several key designs of transformer in the building blocks to capture long-range pixel interactions. Other recent works~\cite{chen2205activating, chen2022cross, li2023efficient, zhang2022accurate, xiao2022stochastic,tsai2022stripformer,chen2023dual} focus on the different window shape or self-attention mechanisms to enlarge the range of toke interactions, but still can not establish truly long-range dependencies across windows. Recently, transformer architecture~\cite{peebles2023scalable, bao2023all, cao2022exploring, hatamizadeh2023diffit, ma2024sit} is also popular in image generation, replacing the regular U-Net backbone of diffusion model in the latent or pixel space, shows much better scaling properties than convolution neural networks.

\section{Method}
\subsection{Motivation}
In image restoration, it is hard to restore clear images from the low-quality input data for that some contents are lost during degradation processes. Thanks to the characteristics of visual image, the missing pattern may exist on other regions of the same one or different images~\cite{freeman2002example}, which could be captured to help restore the high-quality output. CNNs are good learner for different patterns, but the learnable kernels are separated and common, can not provide rich and specific information for some degraded regions. Although researchers explore different structures (VGG-style~\cite{kim2016accurate}, U-shape~\cite{zamir2021multi} and multi-branch~\cite{lim2017enhanced, zhang2020residual}) and attention mechanisms~\cite{lim2017enhanced,wang2020eca,zamir2020learning,zamir2021multi,liu2018non}, but is still hard to establish accurate long-range dependencies. 
\begin{figure*}
	\centering
	\begin{minipage}{0.54\textwidth}
		\centering
		\includegraphics[width=1\linewidth]{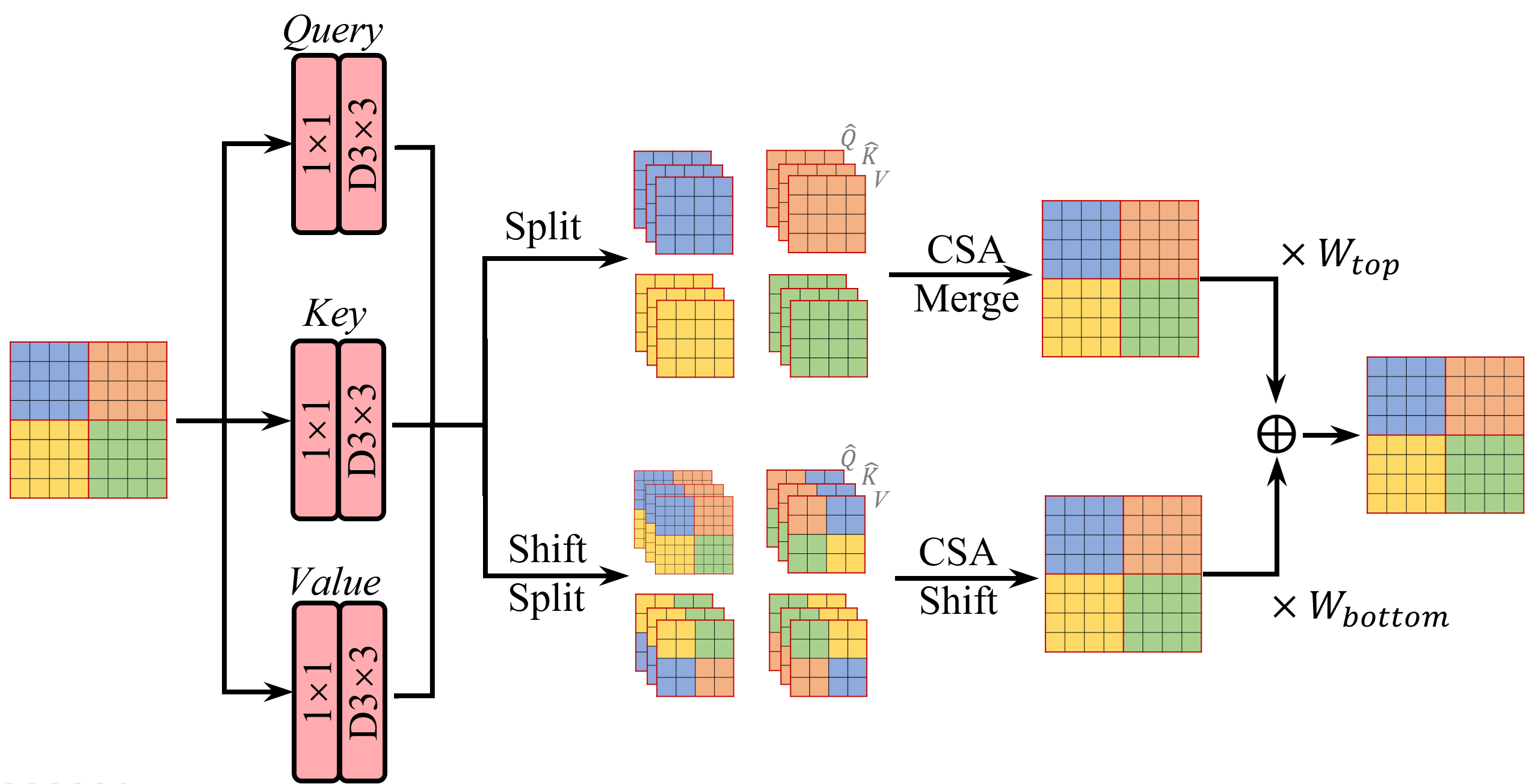}
		\subcaption{Focal Context Self-Attention (FCSA)}
		\label{FCSA}
	\end{minipage}
	\hspace{3mm}
	\begin{minipage}{0.42\textwidth}
		\centering
		\includegraphics[width=1\linewidth]{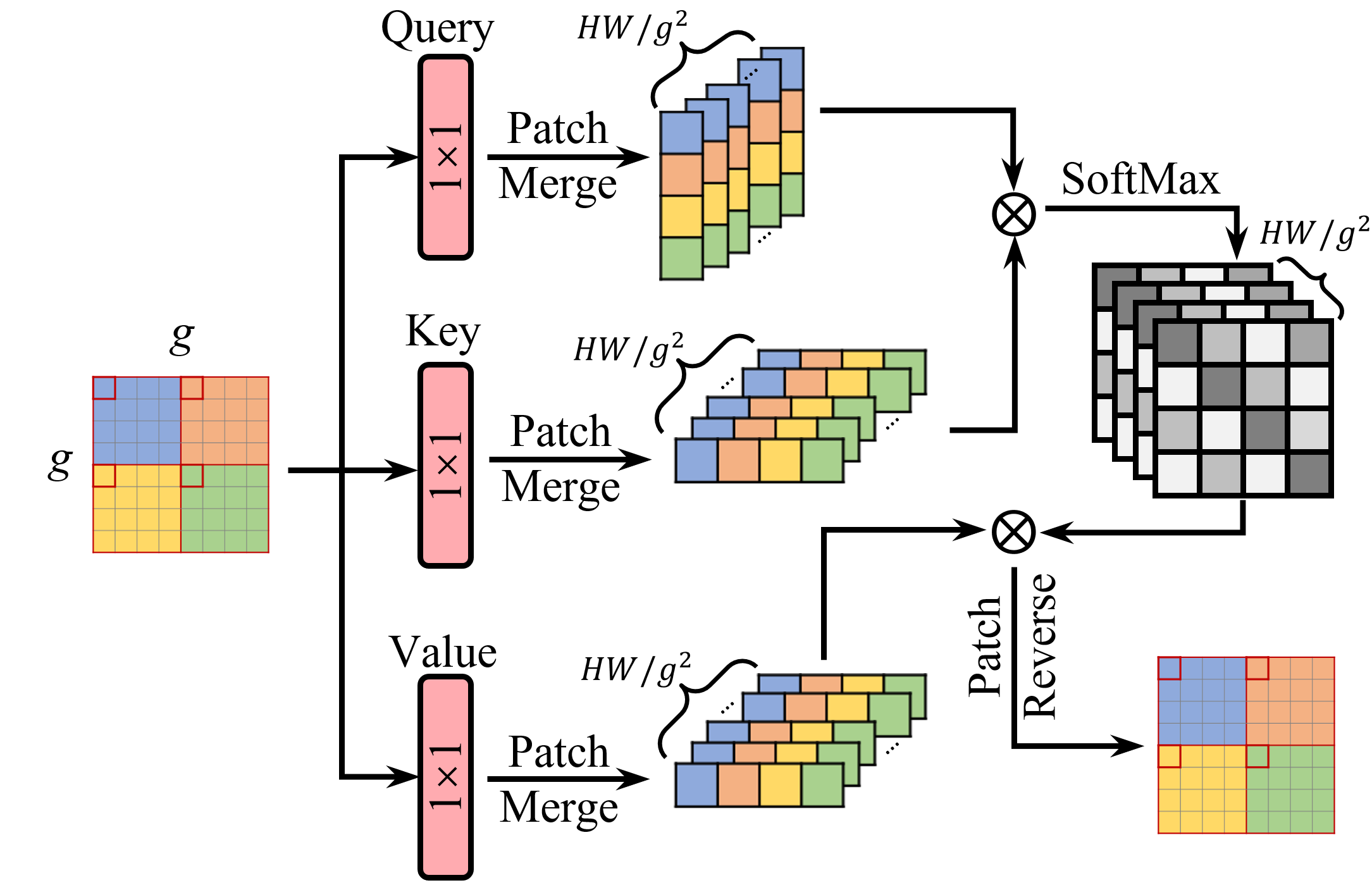}
		\subcaption{Global Grid Self-Attention (GGSA)}
		\label{GGSA}
	\end{minipage}
	\caption{Focal and Global self-attention modules in the proposed IPT-V2. $1\times1$ and $\textrm{D}3\times3$ denote the pointwise convolution and depthwise convolution with kernel size of $3\times3$, respectively. CSA represents the channel self-attention operation as Equation~\ref{Atten_FCSA}. For simplify, the output projection module is hidden in the self-attention architecture, using pointwise convolution as in the previous methods.}
	\label{AttentionBlock}
\end{figure*}

Transformer architecture~\cite{dosovitskiy2020image, chen2021pre} is born to learn the long-range dependencies with self-attention mechanism (SA), and the two commonly used are window-based SA~\cite{liang2021swinir,chen2205activating, chen2022cross, li2023efficient, xiao2022stochastic,tsai2022stripformer} and channel-based SA~\cite{zamir2022restormer}.  Window-based SA calculates the similarities across tokens in the fixed-size window. This method could provide a finer context and larger local information to help restore details and textures than CNNs, but still does not construct complete interactions of different windows. Channel-based SA obtains self-attention metrics by calculating the cosine distance of feature maps in different channels. This method could provide a global but much rough information for that the global interactions are averaged by all the pixels. Furthermore, channel-based SA do not process the local region well, can not help restore details and textures explicitly. To provide accurate global and local context information with less computational complexity, we propose to combine the advantages of channel-based SA and window-based SA, and enhance them in local and global range.  We change the window-based SA with the grid spatial SA, truly construct the long-range dependencies across windows, and partition the feature maps into multiply non-overlapping windows in channel-based SA, provide compact local interactions across channels with less computational overhead.

As shown in Figure~\ref{Motivation}, we analysis the informative pixels that contribute to the restoration of the specific clear region with their Local Attribution Maps (LAM~\cite{gu2021interpreting}). Consistent with our above analysis, CNNs only activate local and small region and SwinIR~\cite{liang2021swinir} captures much more accurate local context but lacks global information. Global pixels are active in Restormer~\cite{zamir2022restormer}, but the interactions are much weak and inaccurate. In contrast, IPT-V2 aims to establish explicit and strong focal and global token interaction to help restore the content and textures for image restoration simultaneously, and the LAM result also proves this intuition. We will describe the details of IPT-V2  in next sections.
\subsection{IPT-V2 Architecture}
As shown in Figure~\ref{IPT-V2}, the overall  architecture of the proposed IPT-V2 is an U-shape encoder-decoder network as~\cite{ronneberger2015u} with 3 times downsampling and upsampling, consists of an input projection layer, multiply focal and global transformer blocks (FGTB) and an output projection layer. Specifically, given a low quality image $I \in \mathbb{R}^{1 \times 3 \times H\times W}$, we firstly employ a convolutional layer with kernel size of $3\times3$ to project the original image to a shallow feature $F_0 \in \mathbb{R}^{1 \times C \times H\times W}$, $H$, $W$ and $C$ denote the height, width and channel of current feature maps. Then the feature maps $F_0$ are extracted by the focal-global transformer blocks of encoder and decoder networks, where the block number of each level gradually increases with downsampling and decreases with upsampling. Following the architecture design of Restormer~\cite{zamir2022restormer}, we also add multiply refinement modules in the last of decoder part. The FGTB contains two sequential transformer layers, employs self-attention in local and global range, respectively, which is defined as :
\begin{equation}
\begin{aligned}
X_l^{'} & = \textrm{FCSA(LN}(X_{l-1})) + X_{l-1}, \\
X_l & = \textrm{Rep-LeFFN(LN}( X_l^{'})) + X_l^{'}, \\
X_{l+1}^{'} & = \textrm{GGSA(LN}(X_{l})) + X_{l}, \\
X_{l+1} & = \textrm{Rep-LeFFN(LN}(X_{l+1}^{'})) + X_{l+1}^{'}
\label{IPT-V2_Equ}
\end{aligned}
\end{equation}
where the FCSA and GGSA represent the focal context self-attention module and global grid self-attention module, LN and Rep-LeFFN denote the layer normalization layer and re-parameterization locally-enhance feed-forward network, we will describe them in the following sections.

For the downsampling, we use a pointwise convolutional layer and pixel-unshuffle operation, which reduce the resolution to half and double the channel number. Compared with convolutional layer with stride of 2, pixel-unshuffle do not loss the feature information in spatial dimension. And for the upsampling, we use pixel-shuffle operation and a convolutional layer, which double the resolution and reduce the channel number to half. To help preserve the original feature information, we add a shortcut connection between the same levels of encoder and decoder except the top level. Then the feature maps of the last transformer block are passed to a convolutional layer with kernel size of $3\times3$, and then output the restored image with the help of a residual connection of original input data. 

\subsection{Focal Context Self-Attention}
Local information is critical to restore the details that is similar with these in the near region. Most previous transformer-based methods~\cite{wang2022uformer,liang2021swinir,li2023efficient,chen2205activating} of image restoration tend to adopt window-based self-attention (WSA) in the spatial dimension, which calculates the attention map in the fixed-size window. Latest methods~\cite{li2023efficient, chen2205activating} usually set the window size of WSA to 16 for better performance, which could provide more accurate  local information over a larger area than CNNs. But the computational complexity of WSA is much large for that is quadratic with the window size, especially on high-resolution images. In contrast, channel self-attention is much computationally efficient, has linear complexity with input resolution. 
Thus we propose a focal context self-attention based on CSA, called FCSA, apply shifted window mechanism into channel self-attention to enhance the local context. Specifically, as shown in Figure~\ref{FCSA}, FCSA contains a shared Query-Key-Value projection module, a parallel channel self-attention module and a SA merge operation. Given a feature map $F \in  \mathbb{R}^{1 \times C \times H\times W}$, FCSA first extracts the \textit{query}, \textit{key} and \textit{value} metrics as Q, K and V ($Q,K,V \in  \mathbb{R}^{1 \times  h \times H \times W\times d}, C = h\times d$) with a sequential connected pointwise convolution layer and depthwise convolution layer. And then we conduct channel self-attention in two branches, the top SA branch is designed to enhance the local context and mutual information among channels and the bottom branch aims to restore the  edge feature of non-overlapping windows. In practical, CB-CSA splits the $Q$, $K$ and $V$ into non-overlapping windows and calculates the channel SA in each window separately, finally get the $F_{top}$ as shown in Equation~\ref{Atten_FCSA}.
\begin{equation}
\begin{aligned}
\hat{Q}, \hat{K} & = \frac{\textrm{w}_tQ}{\Vert \textrm{w}_tQ \Vert_2 }, \frac{\textrm{w}_tK}{\Vert \textrm{w}_tK \Vert_2 },\\
\textrm{Atten}_{FCSA}(\hat{Q}, \hat{K}, V) & = \textrm{SoftMax}(\hat{Q} \cdot \hat{K} / \alpha)V, 
\label{Atten_FCSA}
\end{aligned}
\end{equation}
where the $\Vert \cdot \Vert_2$ calculates the $\mathcal{L}_2$ Norm and $\alpha$ is a learnable temperature parameter to control the magnitude of the dot product. Different from~\cite{zamir2022restormer}, we add an extra token weighting factor $\textrm{w}_t \in \mathbb{R}^{h\times p\times p}$ to learn the spatial importance in the self-attention, $p$ indicates the window size. In the meanwhile, the bottom SA branch shifts the windows of $Q$, $K$ and $V$ as~\cite{liu2021swin}, calculates the channel SA in the shifted window, the formula is the same as the top SA branch in Equation~\ref{Atten_FCSA}. And then the bottom SA branch shifts these windows back and reconstruct the feature maps $F_{bottom}$. Finally, we merges these two branches by weighting the window content of two branches and maintain the edge features of the image. The merging operation is conducted with two learnable parameters ($\textrm{w}_{top}, \textrm{w}_{bottom} \in \mathbb{R}^{h\times p\times p}$) as:
\begin{equation}
\begin{aligned}
F_{out} =\textrm{w}_{top} \times F_{top} + \textrm{w}_{bottom} \times F_{bottom}.
\label{SA_merge}
\end{aligned}
\end{equation}
Benefiting from the shared Query-Key-Value projection and the efficiency of channel self-attention, our proposed FCSA only adds a little extra computational complexity but can significantly enhance the local context information and boost the restoration capability of details compared with the vanilla multi-dconv head transposed attention in~\cite{zamir2022restormer}. Besides,  FCSA also has linear complexity with input resolution, could obtain local context information as window-based self-attention, but at a lower computational overhead.

\subsection{Grid Spatial Self-Attention}

The vanilla self-attention could modeling complete long-range dependencies across tokens, but the complexity is the main drawback, prevent the application for model restoration, especially high resolution. For instance, IPT~\cite{chen2021pre} could only process low resolution input of $48\times48$. Most previous transformer-based methods~\cite{wang2022uformer,liang2021swinir,li2023efficient,chen2205activating} of image restoration tend to adopt shifted window self-attention in the spatial dimension.  However, the shifted mechanism can not provide complete cross-window interaction and global context information to each window. Besides, HAT~\cite{chen2205activating} and GRL~\cite{li2023efficient} designs a parallel channel attention branch to compensate the global information, which is still much rough and inaccurate. To balance the global information and complexity, we propose a global grid self-attention module (GGSA), which computes the spatial self-attention in a global grid, as shown in Figure~\ref{GGSA}. Specifically, we partition the feature map $F \in  \mathbb{R}^{1 \times C \times H\times W}$ into a uniform  $g \times g$ grid in spatial dimension, where each cubby is a fixed-size image patch. Then we calculate the self-attention of pixels with the same position in each cubby, these pixels make up a complete and smaller feature map with much lower resolution compared to the original feature map, thus we could obtain the global self-attention with less computational complexity. The \textit{query}, \textit{key} and \textit{value} are computed by pointwise convolution as Q, K and V ($Q, K, V \in \mathbb{R}^{1 \times HW/g^2 \times h  \times g\times g \times d}$), and the global grid self-attention is formulated as:
\begin{equation}
\begin{aligned}
\textrm{Atten}_{GGSA}(Q, K, V) & = \textrm{SoftMax}(\frac{QK^T}{\sqrt{d}} + B)V,\\
\label{Atten_GGSA}
\vspace{-3mm}
\end{aligned}
\end{equation}
where $B$ represents the relative position embedding calculated as~\cite{liu2021swin}, $h$ and $d$ denotes the head number and hidden dimension of \textit{query}, \textit{key} and \textit{value}. The attention map in GGSA is marked as $A \in \mathbb{R}^{ HW/g^2 \times h \times g^2 \times g^2}$, models the long-range pixel interactions of different cubbies explicitly and provides accurate global context information. 
\begin{figure}[t]
	\centering
	\includegraphics[width=1.0\linewidth]{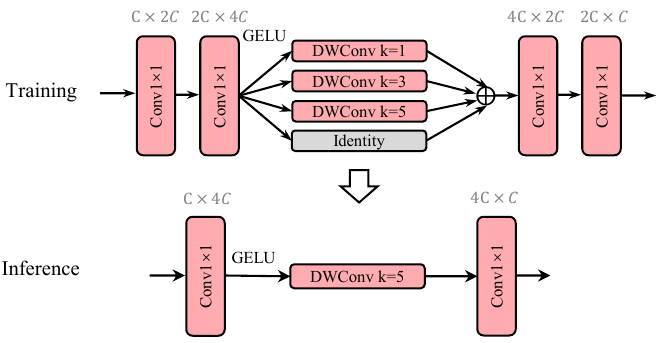}
	\caption{Proposed re-parameterization locally-enhance feed-forward network (Rep-LeFFN).}
	\label{Rep-FFN}
\end{figure}

For the computational complexity, it is well-known that window-based spatial self-attention can reduce the FLOPs compared with vanilla spatial self-attention, from $O(H^2W^2C)$ to $O(p^2HWC)$, but the shifted mechanism is a high-latency operation for the complex memory moving. In our proposed GGSA, the computational complexity is $O(g^2HWC)$ , we can reduce the computational complexity by setting a suitable grid size, and do not need the shift operation to model long-range dependencies.

\subsection{Re-Parameterization LeFFN}
\begin{table*}[t]
\centering
\caption{\textbf{Gaussian image denoising}. The \textcolor{red}{red} and \textcolor{blue}{blue} indicate the best and second best results.}
\label{table:denoising}
\vspace{-2mm}
\setlength{\tabcolsep}{1.5pt}
\scalebox{0.72}{
\begin{tabular}{l | r  r | c c c | c c c | c c c | c c c | c c c | c c c | c c c }
\toprule[0.1em]
\multirow{3}{*}{\textbf{Method}} &  \multicolumn{2}{|c|}{\multirow{3}*{\makecell{ \textbf{FLOPs [G]}\\/ \\ \textbf{Params [M]}}}} &  \multicolumn{12}{c|}{\textbf{Color}} & \multicolumn{9}{c}{\textbf{Grayscale}} \\ \cline{4-24}
&  \multicolumn{2}{r|}{} &
 \multicolumn{3}{c|}{\textbf{CBSD68}~\cite{martin2001database}} & \multicolumn{3}{c|}{\textbf{Kodak24}~\cite{franzen1999kodak}} & \multicolumn{3}{c|}{\textbf{McMaster}~\cite{zhang2011color}} & \multicolumn{3}{c|}{\textbf{Urban100}~\cite{huang2015single}}  & \multicolumn{3}{c|}{\textbf{Set12}~\cite{zhang2017beyond}} & \multicolumn{3}{c|}{\textbf{BSD68}~\cite{martin2001database}} & \multicolumn{3}{c}{\textbf{Urban100}~\cite{huang2015single}} \\
&  \multicolumn{2}{r|}{}
& $\sigma$$=$$15$ & $\sigma$$=$$25$ & $\sigma$$=$$50$ & $\sigma$$=$$15$ & $\sigma$$=$$25$ & $\sigma$$=$$50$ & $\sigma$$=$$15$ & $\sigma$$=$$25$ & $\sigma$$=$$50$ & $\sigma$$=$$15$ & $\sigma$$=$$25$ & $\sigma$$=$$50$ & $\sigma$$=$$15$ & $\sigma$$=$$25$ & $\sigma$$=$$50$ & $\sigma$$=$$15$ & $\sigma$$=$$25$ & $\sigma$$=$$50$ & $\sigma$$=$$15$ & $\sigma$$=$$25$ & $\sigma$$=$$50$ \\ 
\midrule
DnCNN~\cite{kiku2016beyond}	&10.89 & 0.67	&33.90	&31.24	&27.95	&34.60	&32.14	&28.95	&33.45	&31.52	&28.62	&32.98	&30.81	&27.59				&32.86	&30.44	&27.18	&31.73	&29.23	&26.23	&32.64	&29.95	&26.26	\\
RNAN~\cite{zhang2019residual}	& - & 8.96 &-	&-	&28.27	&-	&-	&29.58	&-	&-	&29.72	&-	&-	&29.08				&-	&-	&27.70	&-	&-	&26.48	&-	&-	&27.65	\\
IPT~\cite{chen2021pre}	&276.12&115.33	&-	&-	&28.39	&-	&-	&29.64	&-	&-	&29.98	&-	&-	&29.71				&-	&-	&-	&-	&-	&-	&-	&-	&-	\\
EDT-B~\cite{li2021efficient}	&37.60&11.48	&34.39	&31.76	&28.56	&35.37	&32.94	&29.87	&35.61	&33.34	&30.25	&35.22	&33.07	&30.16				&-	&-	&-	&-	&-	&-	&-	&-	&-	\\
DRUNet~\cite{zhang2021plug}	&35.89 &32.64 &34.30	&31.69	&28.51	&35.31	&32.89	&29.86	&35.40	&33.14	&30.08	&34.81	&32.60	&29.61				&33.25	&30.94	&27.90	&31.91	&29.48	&26.59	&33.44	&31.11	&27.96	\\
SwinIR~\cite{liang2021swinir}	&201.20&11.75	&34.42	&31.78	&28.56	&35.34	&32.89	&29.79	&35.61	&33.20	&30.22	&35.13	&32.90	&29.82				&33.36	&31.01	&27.91	&31.97	&29.50	&26.58	&33.70	&31.30	&27.98	\\
Restormer ~\cite{zamir2022restormer}	&38.83&26.13	&34.40	&31.79	&28.60	&35.47	&33.04	&30.01	&35.61	&33.34	&30.30	&35.13	&32.96	&30.02				&33.42	&31.08	&28.00	&31.96	&29.52	&26.62	&33.79	&31.46	&28.29	\\
ART~\cite{zhang2022accurate} & 286.95 & 16.15 & 34.46 & 31.84 & 28.63 & 35.39 & 32.95 & 29.87 & 35.68 & 33.41 & 30.31 & 35.29 & 33.14 & 30.19 & - & -& -& -& -& -& -& -& - \\
GRL-T~\cite{li2023efficient}	&13.58&0.88	&34.30	&31.66	&28.45	&35.24	&32.78	&29.67	&35.49	&33.18	&30.06	&35.08	&32.84	&29.78				&33.29	&30.92	&27.78	&31.90	&29.43	&26.49	&33.66	&31.23	&27.89	\\
GRL-S~\cite{li2023efficient}	&50.59&3.12	&34.36	&31.72	&28.51	&35.32	&32.88	&29.77	&35.59	&33.29	&30.18	&35.24	&33.07	&30.09				&33.36	&31.02	&27.91	&31.93	&29.47	&26.54	&33.84	&31.49	&28.24	\\
GRL-B~\cite{li2023efficient}	&318.87&19.81	&\textcolor{blue}{34.45}	&\textcolor{blue}{31.82}	&28.62	&\textcolor{blue}{35.43}	&\textcolor{blue}{33.02}	&29.93	&\textcolor{red}{35.73}	&\textcolor{blue}{33.46}	&\textcolor{blue}{30.36}	&\textcolor{red}{35.54}	&\textcolor{red}{33.35}	&\textcolor{blue}{30.46}			&\textcolor{red}{33.47}	&\textcolor{red}{31.12}	& \textcolor{red}{28.03}	&\textcolor{red}{32.00}	&\textcolor{blue}{29.54}	&\textcolor{blue}{26.60}	&\textcolor{red}{34.09}	&\textcolor{blue}{31.80}	&\textcolor{blue}{28.59}	\\	
\midrule
IPT-V2 Small	&14.61&11.75	&34.41	&31.79	&\textcolor{blue}{28.63}	&35.34	&32.92	&29.88	&35.57	&33.32	&30.28	&35.13	&32.98	& 30.09			&33.38	&31.05	&27.97	&\textcolor{blue}{31.98}	&29.53	&26.63	&33.81	&31.51	&28.45	\\
IPT-V2 Base	&45.16&26.49		& \textcolor{red}{34.46}	&\textcolor{red}{31.84}	& \textcolor{red}{28.65}	&	\textcolor{red}{35.44} &\textcolor{red}{33.03}	& \textcolor{red}{29.97} & \textcolor{blue}{35.71}	&\textcolor{red}{33.47}	&	\textcolor{red}{30.42} & \textcolor{blue}{35.39}	 &\textcolor{blue}{33.30}	& \textcolor{red}{30.53}			& \textcolor{blue}{33.44}	& \textcolor{blue}{31.10}	& \textcolor{blue}{28.01}	&\textcolor{red}{32.00}	& \textcolor{red}{29.55}	& \textcolor{red}{26.65}	& \textcolor{blue}{34.04}	& \textcolor{red}{31.84}	& \textcolor{red}{28.83}	\\	
\bottomrule[0.1em]
\end{tabular}}
\end{table*}

\begin{table*}[t]
\begin{center}
\caption{\small \textbf{Real image denoising} on SIDD~\cite{abdelhamed2018high} and DND~\cite{plotz2017benchmarking} datasets.}
\label{realdenoising}
\vspace{-2mm}
\setlength{\tabcolsep}{5.5pt}
\scalebox{0.57}{
\begin{tabular}{c| c| c c c c c c c c c c c c c c c | c  c}
\toprule[0.15em]
\multirow{2}{*}{\textbf{Dataset}} & \multirow{2}{*}{\textbf{Method}}  & BM3D & CBDNet   & RIDNet  & AINDNet  & VDN & SADNet &DANet & CycleISP & MIRNet & DeamNet & MPRNet & DAGL &  Uformer & 
  Restormer & NAF-Net(c32) & IPTV2-S & IPTV2-B\\
 & & \cite{dabov2007image} & \cite{guo2019toward} & \cite{anwar2019real} & \cite{kim2020transfer} & \cite{yue2019variational} &  \cite{chang2020spatial}	 & \cite{yue2020dual} & \cite{zamir2020cycleisp} &  \cite{zamir2020learning} & \cite{ren2021adaptive} & \cite{zamir2021multi} &  \cite{kim2016accurate} &  \cite{wang2022uformer} & \cite{zamir2022restormer} & \cite{chen2022simple} & \multicolumn{2}{c}{(Ours)} \\
\midrule[0.15em]
\textbf{SIDD} & PSNR~$\textcolor{black}{\uparrow}$  &  25.65 & 30.78  &  38.71  &  39.08  & 39.28  & 39.46  & 39.47 & 39.52  & 39.72  & 39.47  & 39.71 & 38.94 & 39.77 & \textcolor{blue}{40.02} & 39.97 & 39.98 &  \textcolor{red}{40.05}\\
~\cite{abdelhamed2018high}  & SSIM~$\textcolor{black}{\uparrow}$  &  0.685 & 0.801  &  0.951  &  0.954  & 0.956  & 0.957  & 0.957 & 0.957  & 0.959  & 0.957  & 0.958 & 0.953 & 0.959 & \textcolor{blue}{0.960}  & \textcolor{blue}{0.960} & \textcolor{blue}{0.960} &  \textcolor{red}{0.961}\\
\midrule[0.1em]
\textbf{DND} & PSNR~$\textcolor{black}{\uparrow}$  &  34.51 & 38.06  &  39.26  &  39.37  & 39.38  & 39.59  & 39.58 & 39.56  & 39.88  & 39.63  & 39.80 & 39.77 & 39.96 & 40.03  & - & \textcolor{blue}{40.04} &  \textcolor{red}{40.09} \\
~\cite{plotz2017benchmarking}  & SSIM~$\textcolor{black}{\uparrow}$  &  0.851 & 0.942  &  0.953  &  0.951  & 0.952  & 0.952  & 0.955 & 0.956  & 0.956  & 0.953  & 0.954 & \textcolor{blue}{0.956}  & \textcolor{blue}{0.956}  & \textcolor{blue}{0.956}  & - & \textcolor{blue}{0.956} & \textcolor{red}{0.957} \\
\bottomrule
\end{tabular}}
\end{center}\vspace{-1.5em}
\end{table*}

\begin{table*}[t]
	\begin{center}
		\caption{\small \textbf{Single-image motion deblurring} results. The \textcolor{red}{red} and \textcolor{blue}{blue} indicate the best and second best results.}
		\label{motion_deblurring}
		\vspace{-2mm}
		\setlength{\tabcolsep}{4pt}
		\scalebox{0.56}{
			\begin{tabular}{c| c| c c c c c c c c c c c c c c c c c | c}
				\toprule[0.15em]
				\multirow{2}{*}{\textbf{Dataset}} & \multirow{2}{*}{\textbf{Method}}  & Xu \textit{etc}& Kupyn \textit{etc} & Nah \textit{etc} & Kupyn \textit{etc}   & SRN & DBGAN & MT-RNN & DMPHN  & Suin \textit{etc} & SPAIR &Cho \textit{etc} & IPT & MPRNet  & Restormer & NAF-Net & GRL-B & DiffIR & IPT-V2 Base\\
				& &  \cite{xu2013unnatural} & \cite{kupyn2019deblurgan} & \cite{gopro2017} &  \cite{deblurganv2} & \cite{tao2018scale} &  \cite{zhang2020dbgan} & \cite{mtrnn2020} & \cite{dmphn2019}  & \cite{Maitreya2020}  & \cite{purohit2021spatially_spair}  &\cite{cho2021rethinking_mimo}  &  \cite{chen2021pre} & \cite{Zamir_2021_CVPR_mprnet} & \cite{zamir2022restormer} & \cite{chen2022simple} &\cite{li2023efficient}  & \cite{xia2023diffir} &  (Ours) \\
				\midrule[0.15em]
				\multicolumn{2}{c|}{\textbf{Params (M)}} &  & &	- &	- &	- &	- &	- &	- &	- &	- &	- &	115.33 &	15.74 &	26.13 & 67.89 &	 19.81 & 26.00 &	 45.16 \\
				\multicolumn{2}{c|}{\textbf{FLOPs (G)}} &  & &	- &	- &	- &	- &	- &	- &	- &	- &	- &	276.12 &	143.36 &	38.83 &  15.81 &	 318.87 & 51.63 &	 45.16 \\
				\midrule[0.1em]
				\textbf{GoPro} & PSNR~$\textcolor{black}{\uparrow}$  & 21.00 &	28.70 &	29.19 &	29.55 &	30.26 &	31.10 &	31.15 &	31.20 &	31.85 &	32.06 &	32.45 &	32.52 &	32.66 &	32.92 & 33.71 &	 \textcolor{red}{33.93}& 33.20 &	 \textcolor{blue}{33.92}  \\
				~\cite{nah2017deep}  & SSIM~$\textcolor{black}{\uparrow}$  & 0.741  &	0.858 &	0.931 &	0.934 &	0.934 &	0.942 &	0.945 &	0.940 &	0.948 &	0.953 &	0.957 &	- &	0.959 &	0.961 & \textcolor{blue}{0.967} 
					 &	\textcolor{red}{0.968} &	 0.963 & \textcolor{red}{0.968} \\
				\midrule[0.1em]
				\textbf{HIDE} & PSNR~$\textcolor{black}{\uparrow}$ & - &	24.51 &	- &	26.61 &	28.36 & 	28.94 &	29.15 &	29.09 &	29.98 &	30.29 &	29.99 &	- 	& 30.96 &	31.22 & -
					 &	\textcolor{blue}{31.65} &	  31.55 & \textcolor{red}{31.74} \\
				~\cite{shen2019human}  & SSIM~$\textcolor{black}{\uparrow}$ & - &	0.871 &	- &	0.875 & 	0.915 &	0.915 &	0.918 &	0.924 &	0.930 &	0.931 &	0.930 &	- &	0.939 &	\textcolor{red}{0.948} & -
					 &	\textcolor{blue}{0.947} &	\textcolor{blue}{0.947} &   \textcolor{blue}{0.947}\\
				\bottomrule
		\end{tabular}}
	\end{center}\vspace{-1.5em}
\end{table*}

Feed-Forward Network (FFN) is the core component in vision transformer architecture design, performs point-wise channel mixing with the expansion-squeeze transformation. However, previous methods~\cite{li2021localvit, wang2022uformer, wu2021cvt, xiao2022image, yuan2021incorporating} show that vanilla FFN lacks the locality of spatial context in vision tasks and present a Locally-enhance Feed-Forward Network (LeFFN), which inserts a depthwise convolution layer into these two pointwise convolution layers to improve the performance for that the local content is much critical to help recover the details, as shown in Equation~\ref{equ_FFN}.
\begin{equation}
\begin{aligned}
\textrm{FFN}(x) & = \textrm{PW-Conv}(\textrm{GELU}(\textrm{PW-Conv}(x))), \\
\textrm{LeFFN}(x) &   = \textrm{PW-Conv}(\textrm{DW-Conv}(\textrm{GELU}(\textrm{PW-Conv}(x)))),
\label{equ_FFN}
\end{aligned}
\end{equation}
where PW-Conv and DW-Conv represent the pointwise convolution and depthwise convolution, respectively. GELU denotes the Gaussian Error Linear Unit~\cite{hendrycks2016gaussian}, which is an activation function that used in most transformer architectures.
Inspired by~\cite{guo2020expandnets, ding2021repvgg}, we propose to apply the structural re-parameterization technique to the LeFFN for extracting more accurate local feature during training, and maintain the original structure in the inference stage, called Rep-LeFFN as shown in Figure~\ref{Rep-FFN}. We adopt sequential and parallel rep-parameterization for pointwise convolution and depthwise convolution simultaneously in the Rep-FFN.
\begin{equation}
\begin{aligned}
\textrm{RepPW-Conv}(x)   =&  \textrm{PW-Conv}((\textrm{PW-Conv}(x)), \\
\textrm{RepDW-Conv}(x)   =& \textrm{DW-Conv}_{5\times5}(x) +\textrm{DW-Conv}_{3\times3}(x) \\
& + \textrm{DW-Conv}_{1\times1}(x) + x, \\
\textrm{Rep-LeFFN}(x)   =& \textrm{RepPW-Conv}(\textrm{RepDW-Conv}(\\
&\textrm{GELU}(\textrm{RepPW-Conv}(x)))),
\label{RepFFN}
\end{aligned}
\end{equation}
where the subscript of DW-Conv indicates its size of convolution kernel. Different from Rep-VGG~\cite{ding2021repvgg} and RepSR~\cite{wang2022repsr}, we do not use batch normalization (BN) layer in each branch before addition for that BN is much unfriendly to the model restoration tasks, such as reducing range flexibility and bringing artifacts. In the inference stage, the RepPW-Conv and RepDW-Conv could be merged into vanilla PW-Conv and DW-Conv, so that Rep-LeFFN do not bring any extra computational complexity compared to the original LeFFN, but significantly boost the capability of feed-forward network in transformer architecture.

\section{Experiments}
In this section, we conduct extensive experiments to demonstrate the performance of our proposed IPT-V2 on various datasets and benchmarks: denoising, deblurring, deraining and image generation, and compare with previous state-of-the-arts methods. Besides, we also report the ablation studies to show the effectiveness of focal context self-attention, global grid self-attention and Rep-LeFFN in IPT-V2 architecture. We provide IPT-V2 Base and IPT-V2 Small in this paper.
\begin{figure*}[h]
	\centering
	\includegraphics[width=0.9\linewidth]{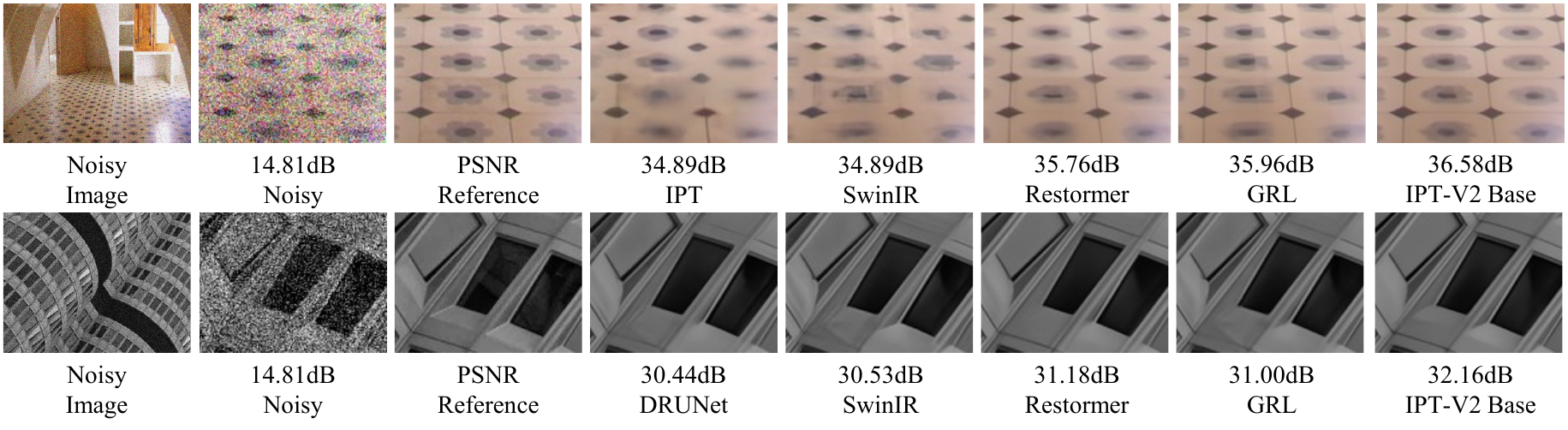}
	\vspace{-3mm}
	\caption{Visual comparison of different methods on image denoising($\sigma=50$, top : Gaussian color, bottom: Gaussian grayscale denoising)}
	\label{visualization}
\end{figure*}
Following the experimental settings of Restormer~\cite{zamir2022restormer}, we adopt the random flips (horizontal or vertical) and rotation on input data, and use progressive learning strategy on mixed-size patches, in which the patch size and batch size pairs are set to [($128^2$,64), ($192^2$,32), ($256^2$,16), ($320^2$,8), ($384^2$,8)] at iterations [0, 116K, 176k, 224K, 270K]. We train IPT-V2 for 300k iters on each dataset, where the initial learning rate is set to 3e-4 and then decay to 1e-6 with Cosine Annealing~\cite{loshchilov2016sgdr}. We adopt AdamW optimizer~\cite{loshchilov2017decoupled} with $\beta=[0.9, 0.999]$ and weight decay of $1e-4$.
\subsection{Image Denoising Results}
\begin{table*}[!t]
\begin{center}
\caption{\textbf{Defocus deblurring} results. \textbf{S:} single-image defocus deblurring. \textbf{D:} dual-pixel defocus deblurring. }
\label{defocus_deblurring}
\vspace{-2mm}
\setlength{\tabcolsep}{8pt}
\scalebox{0.7}{
\begin{tabular}{l | c| c c c c | c c c c | c c c c}
\toprule[0.1em]
\multirow{2}{*}{Method} & \multirow{2}{*}{GFLOPs} & \multicolumn{4}{c|}{\textbf{Indoor Scenes}} & \multicolumn{4}{c|}{\textbf{Outdoor Scenes}} & \multicolumn{4}{c}{\textbf{Combined}} \\ \cline{3-14}
	& &PSNR$\uparrow$ & SSIM$\uparrow$ & MAE$\downarrow$ & LPIPS$\downarrow$			&PSNR$\uparrow$ & SSIM$\uparrow$ & MAE$\downarrow$ & LPIPS$\downarrow$ &PSNR$\uparrow$ & SSIM$\uparrow$ & MAE$\downarrow$ & LPIPS$\downarrow$		\\ \midrule[0.1em]
EBDB$_S$~\cite{karaali2017edge}	& - &25.77 & 0.772 & 0.040 & 0.297				&21.25 & 0.599 & 0.058 & 0.373				&23.45 & 0.683 & 0.049 & 0.336		\\
DMENet$_S$~\cite{lee2019deep}	& - &25.50 & 0.788 & 0.038 & 0.298				&21.43 & 0.644 & 0.063 & 0.397				&23.41 & 0.714 & 0.051 & 0.349		\\
JNB$_S$~\cite{shi2015just}	& - &26.73 & 0.828 & 0.031 & 0.273				&21.10 & 0.608 & 0.064 & 0.355				&23.84 & 0.715 & 0.048 & 0.315		\\
DPDNet$_S$~\cite{abuolaim2020defocus}	& - &26.54 & 0.816 & 0.031 & 0.239				&22.25 & 0.682 & 0.056 & 0.313				&24.34 & 0.747 & 0.044 & 0.277		\\
KPAC$_S$~\cite{son2021single}	& - &27.97 & 0.852 & 0.026 & 0.182				&22.62 & 0.701 & 0.053 & 0.269				&25.22 & 0.774 & 0.040 & 0.227		\\
IFAN$_S$~\cite{lee2021iterative}	& - &28.11 & 0.861 & 0.026 & 0.179				&22.76 & 0.720 & 0.052 & 0.254				&25.37 & 0.789 & 0.039 & 0.217		\\
Restormer$_S$~\cite{zamir2022restormer}	& 38.83 &28.87 & 0.882 & \textcolor{blue}{0.025} & 0.145			&23.24 & 0.743 & \textcolor{blue}{0.050} & 0.209			&25.98 & 0.811 & \textcolor{blue}{0.038} & \textcolor{blue}{0.178}	\\
GRL$_S$-B~\cite{li2023efficient}	& 318.87 &\textcolor{red}{29.06} & \textcolor{red}{0.886} & \textcolor{red}{0.024} & \textcolor{red}{0.139}			&\textcolor{blue}{23.45} & \textcolor{red}{0.761} & \textcolor{red}{0.049} & \textcolor{blue}{0.196}			&\textcolor{red}{26.18} & \textcolor{red}{0.822} & \textcolor{red}{0.037} & \textcolor{red}{0.168}		\\
IPT-V2$_S$ Small	& 14.61 &	28.72 & 0.879 & \textcolor{blue}{0.025} & 0.152 &	23.07 & \textcolor{blue}{0.748} & 0.051 & 0.219 &	25.82 & \textcolor{blue}{0.812} & \textcolor{blue}{0.038} & 0.187 \\
IPT-V2$_S$ Base	& 45.16 & \textcolor{blue}{28.88} & \textcolor{blue}{0.885} & \textcolor{red}{0.024} & \textcolor{blue}{0.141} & \textcolor{red}{23.53} & \textcolor{red}{0.761} & \textcolor{red}{0.049} & \textcolor{red}{0.194}  &  \textcolor{blue}{26.13} & \textcolor{red}{0.822} & \textcolor{red}{0.037} & \textcolor{red}{0.168} \\
 \midrule[0.1em]
DPDNet$_D$~\cite{abuolaim2020defocus}	& -	&27.48 & 0.849 & 0.029 & 0.189				&22.90 & 0.726 & 0.052 & 0.255				&25.13 & 0.786 & 0.041 & 0.223		\\
RDPD$_D$~\cite{abuolaim2021learning}	& -	&28.10 & 0.843 & 0.027 & 0.210				&22.82 & 0.704 & 0.053 & 0.298				&25.39 & 0.772 & 0.040 & 0.255		\\
Uformer$_D$~\cite{wang2022uformer}	&22.36	&28.23 & 0.860 & 0.026 & 0.199				&23.10 & 0.728 & 0.051 & 0.285				&25.65 & 0.795 & 0.039 & 0.243		\\
IFAN$_D$~\cite{lee2021iterative}	& -	&28.66 & 0.868 & 0.025 & 0.172				&23.46 & 0.743 & 0.049 & 0.240				&25.99 & 0.804 & 0.037 & 0.207		\\
Restormer$_D$~\cite{zamir2022restormer}	& 38.83	&29.48 & 0.895 & \textcolor{blue}{0.023} & 0.134			&23.97 & 0.773 & 0.047 & 0.175				&26.66 & 0.833 & \textcolor{blue}{0.035} & 0.155		\\
GRL$_D$-B~\cite{li2023efficient}	& 318.87	&\textcolor{blue}{29.83} & \textcolor{red}{0.903} & \textcolor{red}{0.022} & \textcolor{red}{0.114}				&\textcolor{blue}{24.39} & \textcolor{blue}{0.795} & \textcolor{blue}{0.045} & \textcolor{red}{0.150}				&\textcolor{blue}{27.04} & \textcolor{blue}{0.847} & \textcolor{red}{0.034} & \textcolor{red}{0.133}	\\
IPT-V2$_D$ Small	& 14.61	& 29.43 & 0.891 & \textcolor{blue}{0.023} & 0.133				&		24.23 & 0.787 & 0.046 & 0.175		& 26.76 & 0.838 & \textcolor{blue}{0.035} & 0.155	\\
IPT-V2$_D$ Base		& 45.16 &\textcolor{red}{29.90} & \textcolor{blue}{0.901} & \textcolor{red}{0.022} & \textcolor{blue}{0.122}				&\textcolor{red}{24.62} & \textcolor{red}{0.799} & \textcolor{red}{0.044} & \textcolor{blue}{0.157}			&\textcolor{red}{27.19} & \textcolor{red}{0.849} & \textcolor{red}{0.034} & \textcolor{blue}{0.140}	\\
\bottomrule[0.1em]
\end{tabular}}
\end{center}
\end{table*}

\begin{table*}[t]
	\begin{center}
		\caption{\small \textbf{Image deraining} results. The \textcolor{red}{red} and \textcolor{blue}{blue} indicate the best and second best results.}
		\label{deraining}
		\vspace{-2mm}
		\setlength{\tabcolsep}{5pt}
		\scalebox{0.7}{
			\begin{tabular}{c| c| c c c c c c c c c c | c c}
				\toprule[0.15em]
				\multirow{2}{*}{\textbf{Dataset}} & \multirow{2}{*}{\textbf{Method}}  & DerainNet & SEMI & DIDMDN   & UMRL  & RESCAN & PreNet & MPRNet & MSPFN & SPAIR & Restormer  & IPT-V2 Small & IPT-V2 Base\\
				& & \cite{fu2017clearing} & \cite{wei2019semi} & \cite{zhang2018density} & \cite{yasarla2019uncertainty}  & \cite{li2018recurrent} & \cite{ren2019progressive} 	 & \cite{Zamir_2021_CVPR_mprnet} & \cite{mspfn2020}   &  \cite{purohit2021spatially_spair} & \cite{zamir2022restormer} &  \multicolumn{2}{c}{(Ours)} \\
				\midrule[0.15em]
				\textbf{Rain100H} & PSNR~$\textcolor{black}{\uparrow}$  &  14.92 &	16.56 &	17.35 &	26.01 &	26.36 &	26.77 &	28.66 &	30.41 &	30.95 & 31.46 & \textcolor{blue}{31.54} & \textcolor{red}{31.73}  \\
				~\cite{yang2017deep}  & SSIM~$\textcolor{black}{\uparrow}$  &  0.592 &	0.486 &	0.524 &	0.832 &	0.786 &	0.858 &	0.860 &	0.890 &	0.892 &	0.904 & \textcolor{blue}{0.907} & \textcolor{red}{0.911} \\
				\midrule[0.1em]
				\textbf{Rain100L} & PSNR~$\textcolor{black}{\uparrow}$  &  27.03 &	25.03 &	25.23 &	29.18 &	29.80 &	32.44 &	32.40 &	36.40 &	36.93 &	\textcolor{blue}{38.99} & \textcolor{blue}{38.99} & \textcolor{red}{39.23}\\
				~\cite{yang2017deep}  & SSIM~$\textcolor{black}{\uparrow}$  & 0.884 &	0.842 &	0.741 &	0.923 &	0.881 &	0.950 &	0.933 &	0.965 &0.969 	&\textcolor{red}{0.978} & \textcolor{blue}{0.977} & \textcolor{red}{0.978}\\
				\midrule[0.1em]
				\textbf{Test2800} & PSNR~$\textcolor{black}{\uparrow}$  &  24.31 &	24.43 &	28.13 &	29.97 &	31.29 &	31.75 &	32.82 &	33.64 &	33.34 &	\textcolor{blue}{34.18} & 33.93 & \textcolor{red}{34.38}\\
				~\cite{fu2017removing}  & SSIM~$\textcolor{black}{\uparrow}$  &  0.861& 	0.782 &	0.867 &	0.905 &	0.904 &	0.916 &	0.930 &	0.938 &	0.936 &	\textcolor{blue}{0.944} & 0.943 & \textcolor{red}{0.946}  \\
				\bottomrule
		\end{tabular}}
	\end{center}
\end{table*}

\begin{table}[t]
	\begin{center}
		\caption{\small \textbf{Image generation} results on ImageNet 256 $\times$ 256 dataset. CFG represents classifier-free guidance scale and iters denotes the training iterations.}
		\label{imagenet}
		\vspace{-2mm}
		\setlength{\tabcolsep}{5pt}
		\scalebox{0.68}{
		\begin{tabular}{c | c c c | c c c c c }
			\toprule[0.15em]
			Model & CFG & FLOPs & Iters & FID~$\textcolor{black}{\downarrow}$ & IS~$\textcolor{black}{\uparrow}$ & sFID~$\textcolor{black}{\downarrow}$ & Precision~$\textcolor{black}{\uparrow}$ & Recall~$\textcolor{black}{\uparrow}$ \\
			\midrule[0.15em]	
			DiT-S/2 & 1.0 & 6.06 G & 200k & 78.57 & 16.66 & 13.23 & 0.3245 & 0.5089 \\
			IPTV2-S & 1.0 & 6.24 G & 200K & 43.59 & 34.50 & 9.76 & 0.4909 & 0.6095 \\
			\midrule[0.1em]
			DiT-S/2 & 1.0 & 6.06 G & 400k &  66.59 & 20.68 & 11.76 & 0.3728 & 0.5674 \\
			IPTV2-S & 1.0 & 6.24 G & 400K & 32.69 & 48.18 & 8.99 & 0.5394 & 0.6268 \\
			\midrule[0.1em]
			DiT-S/2 & 1.0 & 6.06 G & 600k &  61.32 & 22.85 & 11.36 & 0.3929 & 0.5896 \\
			IPTV2-S & 1.0 & 6.24 G & 600K & 28.70 & 55.69 & 8.82 & 0.5564 & \textcolor{red}{0.6419} \\
			\midrule[0.1em]
			DiT-S/2 & 1.0 & 6.06 G & 800k &  57.76 & 24.43 & 10.77 & 0.4102 & 0.6056 \\
			IPTV2-S & 1.0 & 6.24 G & 800K & 26.27 & 60.87 & 8.57 & 0.5706 & \textcolor{blue}{0.6361} \\
			\midrule[0.1em]
			DiT-S/2 & 1.5 & 6.06 G & 200k & 57.08 & 25.41 & 10.14 & 0.4061 & 0.5084 \\
			IPTV2-S & 1.5 & 6.24 G & 200K & 20.68 & 73.92 & 7.39 & 0.6462 & 0.5278 \\
			\midrule[0.1em]
			DiT-S/2 & 1.5 & 6.06 G & 400k & 43.01 & 35.80 & 8.65 & 0.4826 & 0.5602 \\
			IPTV2-S & 1.5 & 6.24 G & 400K & 11.65 & 114.14 & 6.43 & 0.7231 & 0.5225 \\
			\midrule[0.1em]
			DiT-S/2 & 1.5 & 6.06 G & 600k & 37.00 & 42.18 & 8.24 & 0.5123 & 0.5638 \\
			IPTV2-S & 1.5 & 6.24 G & 600K & \textcolor{blue}{9.05} & \textcolor{blue}{134.66} & \textcolor{blue}{6.25} & \textcolor{blue}{0.7473} & 0.5235 \\
			\midrule[0.1em]
			DiT-S/2 & 1.5 & 6.06 G & 800k & 33.56 & 47.04 & 7.91 & 0.5378 & 0.5696 \\
			IPTV2-S & 1.5 & 6.24 G & 800K & \textcolor{red}{7.71} & \textcolor{red}{147.47} & \textcolor{red}{6.07} & \textcolor{red}{0.7619} & 0.5240 \\
			\bottomrule
		\end{tabular}}
	\end{center}
\end{table}

\begin{table*}[htb]
	\scriptsize
	\caption{Ablation studies of IPT-V2.}
	\vspace{-2mm}
	\begin{subtable}[h]{0.33\textwidth}
	\centering
	\setlength{\tabcolsep}{5pt}
	\scalebox{0.95}{
		\begin{tabular}{l | c | c | c  }
			\toprule[0.1em]
			\textbf{Network} & \textbf{FLOPs(G)} & \textbf{Params(M)}& \textbf{PSNR}\\
			\midrule[0.1em]
			Restormer & 38.83 & 26.11 & 30.02 \\
			+ GGSA  & 42.86 & 26.01 & 30.44  \\
			+ FCSA  & 43.90 & 26.06 & 30.48 \\
			+ Rep-LeFFN & 45.16 & 26.49 & 30.53 \\
			\bottomrule[0.1em]
		\end{tabular}}
	\caption{\small Architecture }
	\label{Architecture}
	\end{subtable}
	\begin{subtable}[h]{0.33\textwidth}
		\centering
		\setlength{\tabcolsep}{5pt}
		\renewcommand\arraystretch{1.2} 
		\scalebox{0.95}{
			\begin{tabular}{l | c | c | c  }
				\toprule[0.1em]
				\textbf{SA module}  & \textbf{FLOPs(G)} & \textbf{Params(M)}& \textbf{PSNR}\\
				\midrule[0.1em]
				All FCSA & 42.17 & 26.64 & 30.10 \\
				All GGSA  & 45.36 & 25.87 &30.48  \\
				FCSA + GGSA & 45.16 & 26.49 & 30.53 \\
				\bottomrule[0.1em]
		\end{tabular}}
	\caption{\small Self-Attention Block }
	\label{SA}
	\end{subtable}
	\begin{subtable}[h]{0.33\textwidth}
		\centering
		\setlength{\tabcolsep}{5pt}
		\scalebox{0.95}{
			\begin{tabular}{l | c | c | c  }
				\toprule[0.1em]
				\textbf{FFN module}  & \textbf{FLOPs(G)} & \textbf{Params(M)}& \textbf{PSNR}\\
				\midrule[0.1em]
				FFN & 42.80 & 25.75 & 30.40 \\
				LeFFN  & 45.16 & 26.49  & 30.47  \\
				GDFN  & 43.90 & 26.06 & 30.48  \\
				Rep-LeFFN & 45.16 & 26.49 & 30.53 \\
				\bottomrule[0.1em]
		\end{tabular}}
	\caption{\small Feed-Forward Network }
	\label{FFN}
	\end{subtable}
	\label{Ablation_studies}
\end{table*}

We first conduct image denoising experiments both on the synthetic datasets with additive white Gaussian noise (BSD68~\cite{martin2001database}, kodak24~\cite{franzen1999kodak}, McMaster~\cite{zhang2011color}, Urban100~\cite{huang2015single} and Set12~\cite{zhang2017beyond})  and real-world denoising datasets (SIDD~\cite{abdelhamed2018high} and DND~\cite{plotz2017benchmarking})

\noindent\textbf{Gaussian Images Denoising.} Table~\ref{table:denoising} shows the performance of existing methods and IPT-V2 on Gaussian color and grayscale image denoising tasks. To compare the computational complexity, we also list the FLOPs and Parameters of different methods in the table. With the similar complexity, IPT-V2 Base model achieves 30.53 dB on Urban100 dataset (Gaussian $\sigma=50$ color denoising), outperforms Restormer~\cite{zamir2022restormer}(30.02 dB) by a large margin, and even exceed GRL-L~\cite{li2023efficient}(30.46 dB, 318G Flops). Besides, our small version IPT-V2 Small also gets 30.11 dB on Urban100 dataset, surpasses Restormer and GRL-S with less FLOPs and parameters. As show in Figure~\ref{denoising}, our proposed IPT-V2 not only gets state-of-the-arts performance, but also get better trade-off of PSNR and FLOPs than existing methods, which indicate the efficiency of IPT-V2 architecture. Figure~\ref{visualization} shows visualization results with different methods for color and grayscale denoising. Our proposed IPT-V2 reconstructs much better textures and details while removing the unpleasant noise.

\noindent\textbf{Real Image Denoising.} We also conduct experiments on real-world image denoising datasets and compare with existing methods, the results are shown in Figure~\ref{realdenoising}. As we can see that, our proposed IPT-V2 base model achieves 40.05 dB and 40.09 dB on SIDD~\cite{abdelhamed2018high} and DND~\cite{plotz2017benchmarking}, respectively, gains improvements of 0.03 dB and 0.06 dB than previous state-of-the-arts method Restormer\cite{zamir2022restormer}. Besides, the small version of IPT-V2 also gets excellent performance with less computational complexity, and even surpasses the Restormer on DND~\cite{plotz2017benchmarking} datasets.
\subsection{Image Deblurring Results}
Then we perform experiments on image deblurring tasks, including single image motion deblurring and defocus deblurring (single and dual tasks). For motion deblurring, we train our models on GoPro~\cite{nah2017deep} datasets and test on GoPro~\cite{nah2017deep} and HIDE~\cite{shen2019human}). For defocus deblurring, we train and evaluate the performance of our method on DPDD~\cite{abuolaim2020defocus}.

\noindent\textbf{Single Image Motion Deblurring.} Table\ref{motion_deblurring} shows the experimental results, we also list the recent advance methods IPT~\cite{chen2021pre}, MPRNet~\cite{Zamir_2021_CVPR_mprnet}, Restormer~\cite{zamir2022restormer} and GRL~\cite{li2023efficient}. For the GoPro, our method achieves the same SSIM metric with the pervious best GRL, and 33.92 dB PSNR (drop by 0.01 dB than GRL) with much less computational complexity. And for the HIDE, IPT-V2 base model shows much better performance, exceeds GRL-B by 0.09 dB.

\noindent\textbf{Defocus deblurring.} As shown in Table~\ref{defocus_deblurring}, we present results on single image defocus deblurring and dual image decofus deblurring. For the single image task, our base model achieves much better results on three scenes than Restormer, whose complexity is similar with our base version architecture. Compared with GRL-B model~\cite{li2023efficient} with 318 GB FLOPs, our method also gets similar results on some metrics. For the dual image task, IPT-V2 base model achieves much better results than GRL-B, gets 0.15 dB improvement on combined metric. Besides, the small version IPT-V2 also obtains 26.76 dB, surpasses Restormer~\cite{zamir2022restormer} with much less computational cost.
\subsection{Image Deraining Results}
To evaluate the performance of our method on deraining task, we get the training data from multiple datasets~\cite{fu2017removing, li2016rain, yang2017deep, zhang2018density, zhang2019image} and test on Test100~\cite{zhang2019image}, Rain100H~\cite{yang2017deep} and Test2800~\cite{fu2017removing}. To make a fair comparison, we also list the results of recent methods, such as MPRNet~\cite{Zamir_2021_CVPR_mprnet}, MSPFN~\cite{mspfn2020}, SPAIR~\cite{purohit2021spatially_spair} and Restormer~\cite{zamir2022restormer}. Figure~\ref{deraining} shows the experimental results of these three datasets, as we can see that, our proposed IPT-V2 base model gets the state-of-the-arts performance. Especially on Test2800 that contains 2800 images with rain, our method gets 0.20 dB improvement than Restormer. And our IPT-V2 small model also gets similar or higher performance compare with Restormer, \textit{e.g.}  Rain100H~\cite{yang2017deep} and Rain100L~\cite{yang2017deep}, it proves the strong capability of IPT-V2 in deraining tasks.
\subsection{Image Generation}
We apply our proposed architecture into image generation task, replacing the ViT backbone with IPT-V2 in the DiT~\cite{peebles2023scalable} training framework. Following the setting of DiT-S/2, we use AdaW optimizer with a constant learning rate of 1e-4, the batch size is set to 256. And We use the horizontal flips to augment the data during training. The final layer is zero-initial and the layers of layer normalization are initialized with AdaLN-Zero. We training the latent IPT-V2 on ImageNet-256 datasets~\cite{krizhevsky2012imagenet}, and using 250 DDPM sampling steps in the inference. We measure the performance of image generation with FID~\cite{heusel2017gans}, IS~\cite{salimans2016improved}, sFID~\cite{nash2021generating}, precision~\cite{kynkaanniemi2019improved} and recall~\cite{kynkaanniemi2019improved}. All the metrices are evaluated with 50k samples for a fair comparisons. As shown in Table~\ref{imagenet}, we present the results of IPTV2-S and DiT-S/2 with different training stages (200k, 400k, 600k and 800k) and classifier-free guidance scale (1.0 and 1.5). The FLOPs are measured with one step in the latent space, the input size are $1\times4\times32\times32$. With different training stages and guidance scales, IPTV2-S significantly improve the performance of image generation, outperform DiT-S/2 by a large margin with similar computational cost. And IPT-V2 gets FID of 7.71 with only 800k training iterations, which shows great property of our proposed transformer architecture.

\subsection{Ablation Studies}
To demonstrate the effect of components proposed in IPT-V2, we conduct experiments on Gaussian color image denoising task ($\sigma=50$) and evaluate the performance on Urban100 dataset, the experimental results are shown in Table~\ref{Ablation_studies}. More ablation studies please refer to the appendix.

\noindent\textbf{Ablation on Architecture.} We conduct the experiments by starting with Restormer, and then add the GGSA, FCSA and Rep-LeFFN gradually. Table~\ref{Architecture} shows that when replacing the self-attention block with our proposed GGSA, we get 30.44 dB, achieves 0.42 dB improvement. And then we further add the FCSA into the network, this component boost the accuracy by 0.08 dB. Finally, we replace the FFN with Rep-LeFFN, which is actually IPT-V2, gets 30.53 dB and achieves the state-of-the-arts result. Above experiments indicate that the joint application of our proposed GGSA, FCSA and Rep-LeFFN truly improves the performance.

\noindent\textbf{Ablation on Self-Attention Block} As shown in Figure~\ref{SA}, we discuss the effect of different self-attention blocks. When replacing all the self-attention with FCSA and GGSA, they get 30.10 dB and 30.48 dB. Nevertheless, when changing with the sequential connected FCSA and GGSA, we get 30.53 dB and obtain 0.05 dB improvement with less model complexity compared to all GGSA.

\noindent\textbf{Ablation on Feed-Forward Network}
We replace the different feed-forward networks in our proposed IPT-V2 such as vanilla feed-forward network (FFN)~\cite{dosovitskiy2020image}, locally enhance feed-forward network (LeFFN)~\cite{li2021localvit, wang2022uformer, wu2021cvt, xiao2022image, yuan2021incorporating} and gated-dconv feed-forward network (GDFN)~\cite{zamir2022restormer}. Table~\ref{FFN} shows that when using the LeFFN and GDFN, they get 0.07 dB and 0.08 dB PSNR improvement compared with vanilla FFN, and only increase a few computational overhead. When replacing with our proposed Rep-LeFFN, the complexity is as the same as LeFFN, but we get 0.06 dB improvement and achieves 30.53 dB on Urban100 dataset, which demonstrates the efficiency of Rep-LeFFN.

\section{Conclusion}
In this paper, we present a novel and efficient image processing transformer architecture, named IPT-V2, could construct accurate local and global token interactions simultaneously. 
With the focal context self-attention (FCSA) and global grid self-attention (GGSA), we design a basic module focal and global transformer block for IPT-V2. FCSA introduces shifted window mechanism to channel self-attention for enhancing the local context, and GGSA calculates the spatial self-attention in a global grid, which aggregates the global information and models long-range dependencies. Besides, we apply the structural re-parameterization technique into Locally enhance feed-forward network to further improve the capability of IPT-V2. Extensive experiments on various datasets and benchmarks demonstrate that IPT-V2 achieves the state-of-the-arts performance while keeping computational efficiency.

{
    \small
    \bibliographystyle{ieeenat_fullname}
    \bibliography{main}
}

\clearpage
\newpage

\appendix

\section{More Ablation Studies}
\noindent\textbf{Ablation on Window Size and Grid Size.}
The influence of window size in FCSA and grid size in GGSA are presented in Table~\ref{ablation:gs_ws}, these experiments are conducted with IPT-V2 small model. As we can see that, the window size in FCSA does not affect the computational overhead of IPT-V2, and when grid size is 8 or 4, larger the window size, better the performance. But when equipping with grid size of 16, IPT-V2 with window size of 32 obtains better results. And for grid size of GGSA, we can get better performance with larger grid size, while the FLOPs gets larger. To balance the complexity and accuracy, we set window size and grid size and to 32 and 8 for small version IPT-V2, and set window size and patch size and to 32 and 16 for base version.
\begin{table}[h]
	\centering
	\caption{Ablation study on window size (ws) and grid size (gs).}
	\label{ablation:gs_ws}
	\setlength{\tabcolsep}{4pt}
	\scalebox{0.80}{
		\begin{tabular}{c | c | c c | c c c c}
			\toprule[0.1em]
			\multirow{2}{*}{gs} & \multirow{2}{*}{ws} & \multicolumn{2}{c|}{\multirow{2}*{\makecell{ FLOPs [G]\\ Params [M]}}} & CBSD68 & Kodak24 & McMaster & Urban100 \\
			& & & & \cite{martin2001database} & \cite{franzen1999kodak} & \cite{zhang2011color} & \cite{huang2015single} \\
			\midrule
			\multirow{3}{*}{16} & 16 & 16.85 & 11.76 & 28.62 & 29.92 & 30.33 & 30.26 \\
			& 32 & 16.85 & 11.76 & \textbf{28.62} & \textbf{29.92} & \textbf{30.33} & \textbf{30.28} \\
			& 64 & 16.85 & 11.76 & 28.61 & 29.92 & 30.32 & 30.27 \\
			\midrule
			\multirow{3}{*}{8} & 16 & 14.61 & 11.76 & 28.60 & 29.88 & 30.28 & 30.09 \\
			& 32 & 14.61 & 11.76 & 28.60 & 29.88 & 30.28 & 30.11 \\
			& 64 & 14.61 & 11.76 & 28.60 & 29.88 & 30.29 & 30.13 \\
			\midrule
			\multirow{3}{*}{4} & 16 & 14.05 & 11.76 & 28.58 & 29.85 & 30.23 & 29.94\\
			& 32 & 14.05 & 11.76 & 28.58 & 29.85 & 30.25 & 29.98\\
			& 64 & 14.05 & 11.76 & 28.59 & 29.86 & 30.25 & 30.01\\	
			\bottomrule[0.1em]
	\end{tabular}}
\end{table}

\begin{figure*}[t]
	\centering
	\includegraphics[width=1.0\linewidth]{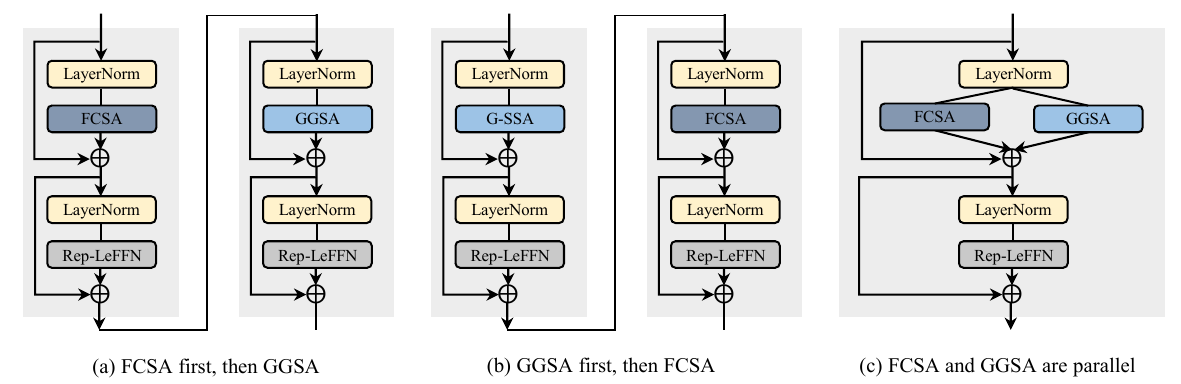}
	\caption{Different orders of FCSA module and G-SSA module.}
	\label{connection}
\end{figure*}

\noindent\textbf{Ablation on Order of FCSA and GGSA.}
We also explore different orders of FCSA and GGSA. As shown in Figure~\ref{connection}, the left is actually IPT-V2, then we get the middle one by switching the order of FCSA and GGSA, and the right represents that we place FCSA and GGSA in parallel and merge their output by a pointwise convolution layer. Table~\ref{ablation:connection} shows that the sequential order of FCSA and GGSA had little effect on the results. And the result of parallel connection is significantly lower than the sequential one, but takes more computational overhead, which indicates that the simple sequential connection of FCSA and GGSA is more efficient to aggregate the local and global information for image restoration.

\begin{table}[h]
	\centering
	\caption{Ablation study on different orders of FCSA and GGSA. $\rightarrow$ and $\vert\vert$ represent the sequential and parallel order in the network.}
	\label{ablation:connection}
	\setlength{\tabcolsep}{5pt}
	\scalebox{0.80}{
		\begin{tabular}{ c | c c c}
			\toprule[0.1em]
			& \ FCSA$\rightarrow$GGSA & GGSA$\rightarrow$FCSA & FCSA $\vert\vert$ GGSA \\
			\midrule
			FLOPs [G] & 45.16 & 45.16 & 47.72\\
			Params [M] & 26.49 & 26.49 & 29.81\\
			\midrule
			CBSD68~\cite{martin2001database} & 28.65 & 28.65 & 28.64\\
			Kodak24~\cite{franzen1999kodak} & 29.97 & 29.97 & 28.64\\ 
			McMaster~\cite{zhang2011color} & 30.42 & 30.41 & 30.39\\
			Urban100~\cite{huang2015single} & 30.53 & 30.52 & 30.45\\	
			\bottomrule[0.1em]
	\end{tabular}}
\end{table}

\section{More Experimental Results}
Table~\ref{exp_setting} shows the architecture details of IPT-V2. To further demonstrate the scalability of our proposed architecture with accurate focal and global self-attention mechanisms, we design an IPT-V2 Base+ model that adds 2 transformer layers to the baseline architecture at each stage. Compared with IPT-V2 Base, Base+ model increases 39.5\% FLOPs and 24.7\% parameters, while the computational overhead is still much lower than previous state-of-the-arts methods GRL-B~\cite{li2023efficient}, SwinIR~\cite{liang2021swinir} and IPT~\cite{chen2021pre}.

\begin{table}[h]
	\footnotesize
	\setlength{\tabcolsep}{1.8mm}
	\renewcommand\arraystretch{1.1}
	\centering
	\caption{Experimental settings of IPT-V2 architecture with different model sizes.  Channel represents the initial channel number. Encoder-Decoder and Refinement denote the transformer block number of encoder, decoder and refinement module. And the window size and patch size are the settings of focal context self-attention and global grid self-attention.}
	\begin{tabular}{l | c  c  c }
		\toprule[0.1em]
		Architecture & IPT-V2 Small & IPT-V2 Base & IPT-V2 Base+ \\
		\midrule
		FLOPs [G] & 14.61 & 45.16 & 63.01 \\
		Params [M] & 11.75 & 26.49 & 33.04 \\
		Channel & 32 & 48 & 48 \\
		Encoder-Decoder & [4,6,6,8] & [4,6,6,8] & [6,8,8,10] \\
		Refinement & 0 & 4 & 6 \\
		Window size & 32 & 32 & 32 \\
		Grid size & 8 & 16 & 16\\
		\bottomrule[0.1em]
	\end{tabular}
	\label{exp_setting}
\end{table}

We list the results of the largest models for different methods in Table~\ref{ablation:denoising}. 
As we can see that, our proposed IPT-V2 Base+ obtains the best performance on various test benchmarks by a large margin. Compared with GRL-B, we get 0.14 dB PSNR improvement on color Urban100 and 0.31 dB PSNR improvement on grayscale Urban100, while the FLOPs is reduce by $5\times$.  

\section{Visualization}
\begin{figure*}[t]
	\centering
	\includegraphics[width=1.0\linewidth]{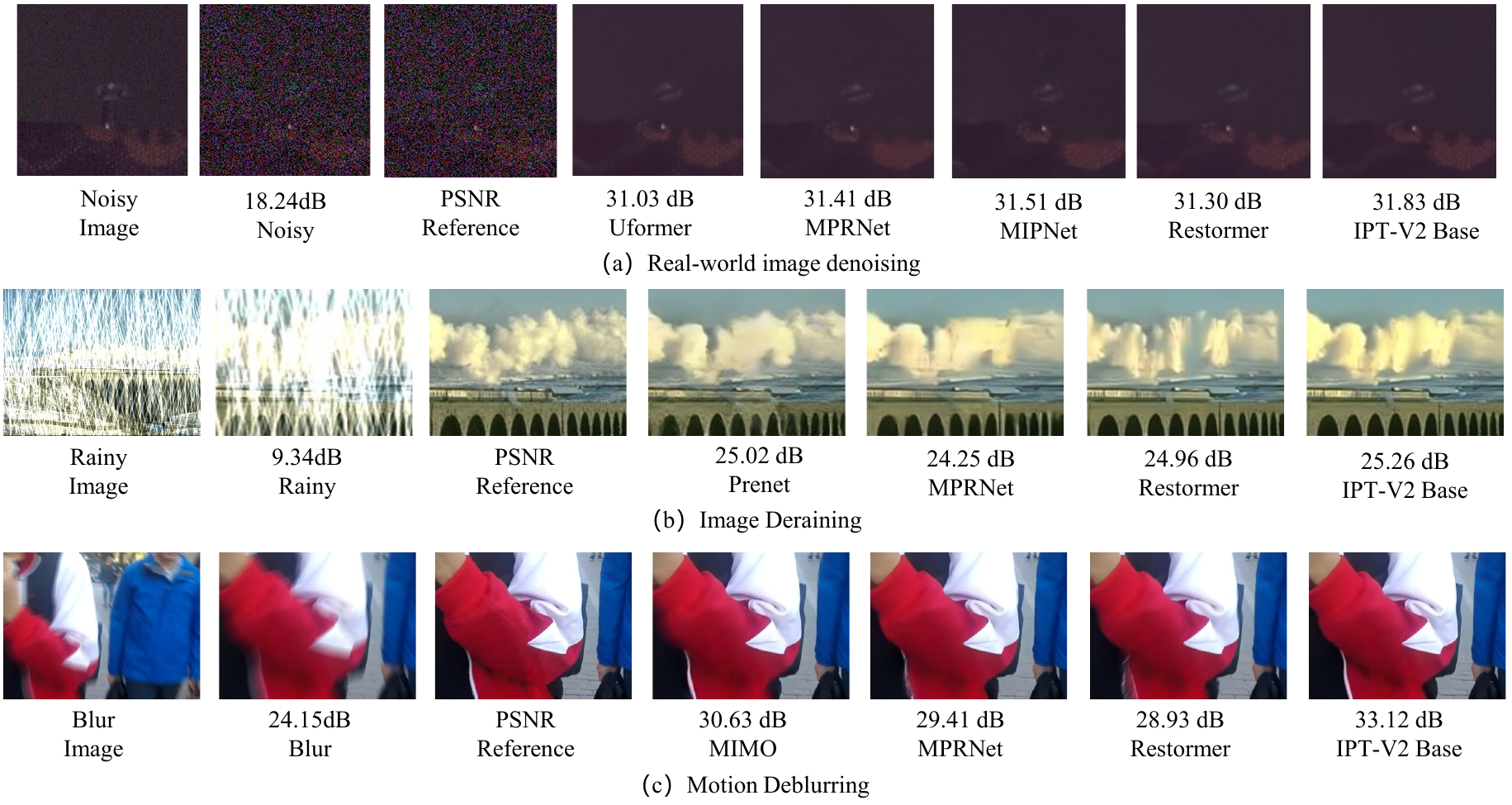}
	\vspace{-5mm}
	\caption{Visual comparison of different methods on real-world image denoising, deraining and motion deblurring tasks. The metrics below are PSNR(dB).}
	\label{visualization_appendix}
\end{figure*}

\begin{figure*}[t]
	\centering
	\includegraphics[width=1.0\linewidth]{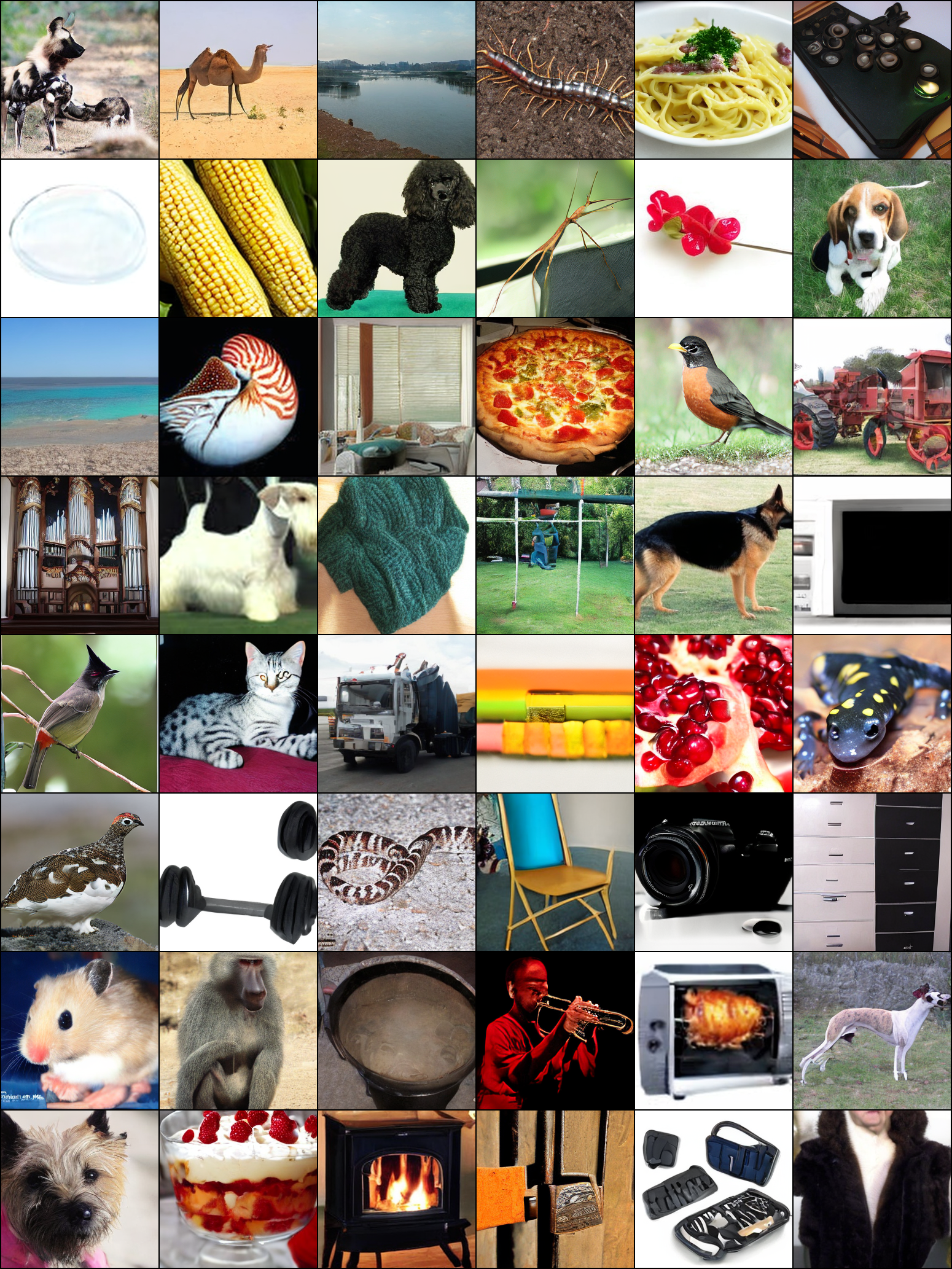}
	\vspace{-5mm}
	\caption{Visualization of uncurated generated 256$\times$×256 images by latent IPT-V2 model. Images are randomly sampled. Classifier-free guidance scale = 4.0.}
	\label{visualization_generation}
\end{figure*}

Figure~\ref{visualization_appendix} shows the visualization of different methods on real-world denoising, deraining and motion deblurring tasks. For reference, we also list the corresponding low-quality images and high-quality images. As we can see that, our proposed IPT-V2 restores much better output, which provides more texture and details. For image generation, we visualize the sample quality in Figure~\ref{visualization_generation}, where we predict the samples at 800k iterations with class-free guidance scale of 4.0.

\begin{table*}[t]
\centering
\caption{\textbf{Gaussian image denoising ($\sigma=50$)} of the largest models for different methods.}
\label{ablation:denoising}
\vspace{-2mm}
\setlength{\tabcolsep}{3pt}
\scalebox{0.8}{
\begin{tabular}{l | r | r | c c c c | c c c }
\toprule[0.1em]
\multirow{2}{*}{\textbf{Method}} &  \multirow{2}*{\textbf{FLOPs [G]}} &  \multirow{2}*{\textbf{Params [M]}} & \multicolumn{4}{c|}{\textbf{Color}} & \multicolumn{3}{c}{\textbf{Grayscale}} \\ \cline{4-10}
&  & &
 \textbf{CBSD68}~\cite{martin2001database} & \textbf{Kodak24}~\cite{franzen1999kodak} & \textbf{McMaster}~\cite{zhang2011color} & \textbf{Urban100}~\cite{huang2015single}  & \textbf{Set12}~\cite{zhang2017beyond} & \textbf{BSD68}~\cite{martin2001database} & \textbf{Urban100}~\cite{huang2015single} \\
\midrule
DnCNN~\cite{kiku2016beyond}	&10.89 & 0.67	&27.95	&28.95	&28.62	&27.59	&27.18	&26.23	&26.26	\\
RNAN~\cite{zhang2019residual}	& - & 8.96 &28.27 &29.58	&29.72 &29.08	&27.70 &26.48	&27.65	\\
IPT~\cite{chen2021pre}	&276.12&115.33	&28.39 &29.64	&29.98	&29.71		&-	&-	&-	\\
EDT-B~\cite{li2021efficient} &37.60&11.48	&28.56	&29.87	&30.25	&30.16	&-	&-	&-		\\
DRUNet~\cite{zhang2021plug}	&35.89 &32.64	&28.51	&29.86	&30.08	&29.61	&27.90	&26.59	&27.96	\\
SwinIR~\cite{liang2021swinir}	&201.20&11.75	&28.56	&29.79	&30.22	&29.82	&27.91	&26.58	&27.98	\\
Restormer ~\cite{zamir2022restormer} &38.83&26.13	&28.60	&30.01	&30.30	&30.02	&28.00	&26.62	&28.29	\\
ART~\cite{zhang2022accurate} & 286.95 & 16.15  & 28.63  & 29.87  & 30.31 & 30.19 & -&  -&  - \\
GRL-B~\cite{li2023efficient}	&318.87&19.81	&28.62	&29.93	&30.36	&30.46	& 28.03	& 26.60	& 28.59	\\	
\midrule
IPT-V2 Base+	&63.01 &33.04	& \textbf{28.65}	& \textbf{30.00} &	\textbf{30.44} &  \textbf{30.60}		& \textbf{28.03}	&	\textbf{26.65}	&	\textbf{28.90}\\	
\bottomrule[0.1em]
\end{tabular}}
\end{table*}

\end{document}